\documentclass{article} 
\usepackage{iclr2023_conference,times}

\usepackage{amsmath,amsfonts,bm}

\def\eqref#1{equation~\ref{#1}}

\def\1{\bm{1}}

\DeclareMathAlphabet{\mathsfit}{\encodingdefault}{\sfdefault}{m}{sl}
\SetMathAlphabet{\mathsfit}{bold}{\encodingdefault}{\sfdefault}{bx}{n}

\DeclareMathOperator*{\argmax}{arg\,max}
\DeclareMathOperator*{\argmin}{arg\,min}

\usepackage{hyperref}
\usepackage{url}
\usepackage{algorithm}
\usepackage{algpseudocode}
\usepackage{graphicx}
\usepackage{caption}
\usepackage{subcaption}
\usepackage{xfrac}
\usepackage{xcolor}
\usepackage{dsfont} 
\usepackage{enumitem}
\setlist[itemize]{align=parleft,left=0pt..1em}

\definecolor{iceblue}{HTML}{039BE5}
\definecolor{firered}{HTML}{CF1920}

\newtheorem{definition}{Definition}

\newtheorem{lemma}{Lemma}

\newcommand{\JH}[1]{{\color{cyan}[Juyeon: #1]}}
\newcommand{\MW}[1]{{\color{red}[MW: #1]}}

\newcommand{\linebreakand}{%
  \end{@IEEEauthorhalign}
  \hfill\mbox{}\par
  \mbox{}\hfill\begin{@IEEEauthorhalign}
}
\makeatother

\newcommand\blfootnote[1]{%
  \begingroup
  \renewcommand\thefootnote{}\footnote{#1}%
  \addtocounter{footnote}{-1}%
  \endgroup
}

\title{Robust Explanation Constraints \\ for Neural Networks}

\iclrfinalcopy

\author{
    Matthew Wicker$^{*,1}$,\quad \textbf{Juyeon Heo}$^{*,2}$,\quad \textbf{Luca Costabello}$^{3}$,\quad \textbf{Adrian Weller}$^{1,2}$ \\
    \textbf{$^1$} The Alan Turing Institute,\quad 
    \textbf{$^2$} University of Cambridge,\quad 
    \textbf{$^3$}  Accenture Labs\\
}

\begin{document}

\maketitle
\begin{abstract}
Post-hoc explanation methods are used with the intent of providing insights about neural networks and are sometimes said to help engender trust in their outputs. However, popular explanations methods have been found to be fragile to minor perturbations of input features or model parameters. Relying on constraint relaxation techniques from non-convex optimization, we develop a method that upper-bounds the largest change an adversary can make to a gradient-based explanation via bounded manipulation of either the input features or model parameters. By propagating a compact input or parameter set as symbolic intervals through the forwards and backwards computations of the neural network we can formally certify the robustness of gradient-based explanations. Our bounds are differentiable, hence we can incorporate provable explanation robustness into neural network training. Empirically, our method surpasses the robustness provided by previous heuristic approaches. We find that our training method is the only method able to learn neural networks with certificates of explanation robustness across all six datasets tested.\blfootnote{Author email addresses in listed order: mwicker@turing.ac.uk, jh2324@cam.ac.uk, luca.costabello@accenture.com, aweller@turing.co.uk}

\end{abstract}

\section{Introduction}

Providing explanations for automated decisions is a principal way to establish trust in machine learning models. 
In addition to engendering trust with end-users, explanations that reliably highlight key input features 
provide important information to machine learning engineers who may use them to aid in model debugging and monitoring \citep{pinto2019automatic, adebayo2020debugging,bhatt2020external,bhatt2020explainable}. The importance of explanations has led to regulators considering them as a potential requirement for deployed decision-making algorithms \citep{gunning2016explainable, goodman2017european}. Unfortunately, deep learning models can return significantly different explanations for nearly identical inputs \citep{dombrowski2019explanations} which erodes user trust in the underlying model and leads to a pessimistic outlook on the potential of explainability for neural networks \citep{rudin2019stop}.
Developing models that provide robust explanations, i.e., that provide similar explanations for similar inputs, is vital to ensuring that explainability methods have beneficial insights \citep{lipton2018mythos}. 

Current works evaluate the robustness of explanations in an adversarial setting by finding minor manipulations to the deep learning pipeline which cause the worst-case (e.g., largest) changes to the explanation \citep{dombrowski2019explanations, heo2019fooling}. It has been shown that imperceptible changes to an input can fool explanation methods into placing importance on arbitrary features \citep{dombrowski2019explanations, ghorbani2019interpretation} and that model parameter can be manipulated to globally corrode which features are highlighted by explanation methods \citep{heo2019fooling}. While practices for improving explanation robustness focus on heuristic methods to avoid current attacks \citep{dombrowski2022towards, wang2020smoothed, chen2019robust}, it is well-known that adversaries can develop more sophisticated attacks to evade these robustness measures \citep{athalye2018obfuscated}. To counter this, our work establishes the first framework for general neural networks which provides a guarantee that for any minor perturbation to a given input's features or to the model's parameters, the change in the explanation is bounded. Our bounds are formulated over the input-gradient of the model which is a common source of information for explanations \citep{sundararajan2017axiomatic, wang2020smoothed}. Our guarantees constitute a formal certificate of robustness for a neural network's explanations at a given input which can provide users, developers, and regulators with a heightened sense of trust. Further, while it is known that explanations of current neural networks are not robust, the differentiable nature of our method allows us to incorporate provable explanation robustness as a constraint at training time, yielding models with significantly heightened explanation robustness.

Formally, our framework abstracts all possible manipulations to an input's features or model's parameters into a hyper-rectangle, a common abstraction in the robustness literature \citep{mirman2018differentiable, gowal2018effectiveness}. 
We extend known symbolic interval analysis techniques in order to propagate hyper-rectangles through both the forwards and backwards pass operations of a neural network. The result of our method is a hyper-rectangle over the space of explanations that is guaranteed to contain every explanation reachable by an adversary who perturbs features or parameters within the specified bounds. We then provide techniques that prove that all explanations in the reachable explanation set are sufficiently similar and thus that no successful adversarial attack exists. Noticing that smaller reachable explanation sets imply more robust predictions, we introduce a novel regularization scheme which allows us minimize the size of of the explanation set during parameter inference. Analogous to state-of-the-art robustness training \citep{gowal2018effectiveness}, this allows users to specify input sets and parameter sets as explainability constraints at train time. Empirically, we test our framework on six datasets of varying complexity from tabular datasets in financial applications to medical imaging datasets. We find that our method outperforms state-of-the-art methods for improving explanation robustness and is the only method that allows for certified explanation robustness even on full-color medical image classification.
We highlight the following contributions:

\begin{itemize}\setlength\itemsep{0.25em}
    \item We instantiate a framework for bounding the largest change to an explanation that is based on the input gradient, therefore certifying that no adversarial explanation exists for a given set of inputs and/or model parameters.
    \item We {\color{black} compute} explicit bounds relying on interval bound propagation, and show that these bounds can be used to regularize neural networks during learning with robust explanations constraints.
    \item Empirically, our framework allows us to certify robustness of explanations and train networks with robust explanations across six different datasets ranging from financial applications to medical image classification.  
\end{itemize}

\section{Related work}

Adversarial examples, inputs that have been imperceptibly modified to induce misclassification, are a well-known vulnerability of neural networks \citep{IntruigingProperties, goodfellow2015explaining, madry2018towards}. A significant amount of research has studied methods for proving that no adversary can change the prediction for a given input and perturbation budget \citep{tjeng2017evaluating, weng2018towards, fazlyab2019safety}. More recently, analogous attacks on gradient-based explanations have been explored \citep{ghorbani2019interpretation, dombrowski2019explanations, heo2019fooling}. In \citep{ghorbani2019interpretation, dombrowski2019explanations} the authors investigate how to maximally perturb the explanation using first-order methods similar to the projected gradient descent attack on predictions \citep{madry2018towards}. In \citep{heo2019fooling, slack2020fooling, lakkaraju2020fool,dimanov2020you} the authors investigate perturbing model parameters rather than inputs with the goal of finding a model that globally produces corrupted explanations while maintaining model performance. This attack on explanations has been extended to a worrying use-case of disguising model bias \citep{dimanov2020you}.

Methods that seek to remedy the lack of robustness work by either modifying the training procedure \citep{dombrowski2022towards, wang2020smoothed} or by modifying the explanation method \citep{pan2020fairwashing, wang2020smoothed, si2021simple}. Methods that modify the model include normalization of the Hessian norm  during training in order to give a penalty on principle curvatures \citep{dombrowski2022towards, wang2020smoothed}. Further work suggests attributional adversarial training, searching for the points that maximize the distance between local gradients \citep{chen2019robust, ivankay2020far, singh2020attributional, Himabindu2020blackbox}. Works that seek to improve explanation in a model agnostic way include smoothing the gradient by adding random noise \citep{smilkov2017smoothgrad, sundararajan2017axiomatic} or by using an ensemble of explanation methods \citep{Laura2020simpledefense}

The above defenses cannot rule out potential success of more sophisticated adversaries and it is well-known that the approaches proposed for improving the robustness of explanations do not work against adaptive adversaries in the case of prediction robustness \citep{athalye2018obfuscated, he2017adversarial}. However, \textit{certification} methods have emerged which provide sound guarantees that even the most powerful adversary cannot fool a models prediction \citep{huang2017safety, gehr2018ai2, wu2020game}. Moreover, these methods have been made differentiable and used for training models with provable prediction robustness and for small networks achieve state-of-the-art robustness results \citep{gowal2018effectiveness, mirman2018differentiable, wicker2021bayesian}. This work extends these certification methods to the problem of robust explanations. By adapting results from symbolic interval analysis of Bayesian neural networks \citep{wicker2020probabilistic, berrada2021make} and applying them to interval analysis of both the forward pass \textit{and} backwards pass we are able to get guarantees on the robustness of gradient-based explanations. {\color{black} Appendix~\ref{appendix:futherlit} summarizes literature on non-gradient-based explanations.}

\section{Background}

We consider neural networks tasked with solving a supervised learning problem where we are given a dataset of $n_{\mathcal{D}}$-many inputs and labels, $\mathcal{D} = \{x^{(i)}, y^{(i)}\}_{i=0}^{n_{\mathcal{D}}-1}$, with inputs $x^{(i)} \in \mathbb{R}^n$, and corresponding target outputs $y^{(i)} \in \mathbb{R}^{m}$ 
either a one-hot class vector for classification or a real-valued vector for regression. We consider a feed forward neural network (NN) as a function $f^\theta:\mathbb{R}^{n}\to\mathbb{R}^m$, parametrised by a vector  $\theta \in \mathbb{R}^{p}$ containing all the weights and biases of the network. 
Given a NN $f^\theta$ composed of $K$ layers, we denote by $f^{\theta,1},...,f^{\theta,K}$ the layers of $f^\theta$ and we have that the parameters are given by the set, $\theta=\big(\{W^{(i)}\}_{i=1}^{K}\big) \cup \big(\{b^{(i)}\}_{i=1}^{K}\big)$, where $W^{(i)}$ and $b^{(i)}$ represent weights and biases of the $i-$th layer of $f^\theta$. Moreover, we take $\sigma$ to be the activation function of the network and $\mathcal{L}$ to be the loss function. Given this, we can write down the forward and backwards pass w.r.t. an input $x$ through a network network as follows: 
\begin{equation*}
\begin{aligned}[c]
    \text{Forward}& \text{ Pass:}\\
    z^{(0)} &= x \\
    \zeta^{(k)} &=   W^{(k)} z^{(k-1)} + b^{(k)}      \\
    z^{(k)} &= \sigma(\zeta^{(k)})   \label{eq:nn3}
\end{aligned}
\qquad \qquad
\begin{aligned}[c]
    \text{Backward}& \text{ Pass:}\\
    d^{(K)} &= \sfrac{\partial \mathcal{L}}{ \partial z^{(K)}} \\
    \delta^{(k-1)} &=  \dfrac{\partial z^{(k)}}{ \partial \zeta^{(k)}} \dfrac{\partial \zeta^{(k)}}{ \partial z^{(k-1)}} \\
    d^{(k-1)} &= d^{(k)} \odot \delta^{(k-1)}  
\end{aligned}
\end{equation*}
%
where $\odot$ represents the Hadamard or element-wise product. We highlight that $d^{(0)} = \partial \mathcal{L}/\partial x$, the gradient of the loss with respect to the input, and will refer to this with the vector $v$, i.e., $v = \nabla_x \mathcal{L}(f^{\theta}(x))$. The input gradient tells us how the NN prediction changes locally around the point $x$ and is therefore a primary source of information for many explanation methods \citep{sundararajan2017axiomatic, zintgraf2017visualizing, smilkov2017smoothgrad}.

\subsection{Attacks on Explanations}

\paragraph{Input Perturbations} Analogous to the standard adversarial attack threat model, targeted input perturbations that attack explanations seek to modify inputs to achieve a particular explanation. We denote an adversary's desired explanation vector as $v^{\text{targ}} \in \mathbb{R}^{n}$. The goal of the adversary is to find an input $x^{\text{adv}}$ such that $h(v^{\text{targ}}, v^{\text{adv}}) \leq \tau$ where $v^{\text{adv}} = \nabla_{x^{\text{adv}}} \mathcal{L}(f^{\theta}(x^{\text{adv}}))$, $h$ is a similarity metric, and $\tau$ is a similarity threshold that determines when the attack is sufficiently close to the target. The structural similarity metric, Pearson correlation coefficient, and mean squared error have all been investigated as forms of $h$ \citep{adebayo2018sanity}. We use the mean squarred error for $h$ throughout this paper {\color{black} and discuss cosine similarity in Appendix~\ref{appendix:cosine}}. 
Moreover, for the untargeted attack scenario, an attacker simply wants to corrupt the prediction as much as possible. Thus no target vector is set and an attack is successful if $h(v, v^{\text{adv}}) \geq \tau$ where $\tau$ is a threshold of dissimilarity between the original explanation and the adversarially perturbed explanation.

\paragraph{Model Perturbations}

Not typically relevant when attacking predictions is the model parameter perturbation setting. In this setting, an attacker wants to perturb the model in a way that maintains predictive performance while obfuscating any understanding that might be gained through explanations. As before, we assume the adversary has a target explanation $v^{\text{targ}}$ and that the adversary is successful if they find a parameter $\theta^{\text{adv}}$ such that $h(v^{\text{targ}}, v^{\text{adv}}) \leq \tau$ where $v^{\text{adv}} = \nabla_{x} \mathcal{L}(f^{\theta^{\text{adv}}}(x))$. The untargeted model attack setting follows the same formulation as in the input attack setting. Model parameter attacks on explanations are usually global and thus one tries to find a single value of $\theta^{\text{adv}}$  that corrupts the model explanation for many inputs at once. This has been investigated in \citep{heo2019fooling, slack2020fooling, lakkaraju2020fool}  and in \citep{pan2020fairwashing, dimanov2020you} where the authors explore modifying models to conceal discriminatory bias. 

\section{Problem Statements}

In this section, we formulate the fragility (the opposite of robustness) of an explanation as the largest difference between the maximum and minimum values of the input gradient for any input inside of the set $T$ and/or any model inside the set $M$. 
We then show how one can use the maximum and minimum input gradients to rule out the existence of successful targeted and untargeted attacks.
We start by formally stating the key assumptions of our method and then proceed to define definitions for robustness. \textit{Assumptions: } the methods stated in the text assume that input sets and weight sets are represented as intervals, we also assume that all activation functions are monotonic, and that all layers are affine transformations e.g., fully connected layers or convolutional layers. We discuss when and how one can relax these assumptions in Appendix~\ref{appendix:assumptions}. Now, we proceed define what it means for a neural network's explanation to be robust around an input $x \in \mathbb{R}^n$.
\begin{definition}\label{def:inputcert} \textbf{\textit{Input-Robust Explanation}} Given a neural network $f^{\theta}$, a test input $x$, a loss function $\mathcal{L}$, a compact input set $T$, and a vector $\mathbf{\delta} \in \mathbb{R}^{n}$,  we say that the explanation of the network $f$ is $\delta$-input-robust iff $\forall i \in [n]$: 
\begin{align}\label{eq:inputrobust}
    \bigg\lVert \min_{x \in T} \dfrac{\partial \mathcal{L}}{ \partial x_i} - \max_{x \in T} \dfrac{\partial \mathcal{L}}{ \partial x_i} \bigg\rVert \leq \delta_i
\end{align}
Moreover we define the set $E_{T} := \{ v'\ |\ \forall i \in [n], \  \min_{x \in T} \partial \mathcal{L}/ \partial x_i \leq v'_i \leq  \max_{x \in T} \partial \mathcal{L} / \partial x_i \}$ to be the reachable set of explanations.
\end{definition}
An intuitive interpretation of this definition is as a guarantee that for the given neural network, input, and loss, there does not exist an adversary that can change the input gradient to be outside of the set $E_{T}$, given that the adversary only perturbs the input by amount specified by $T$. Further, the vector $\delta$ defines the per-feature-dimension fragility (lack of robustness). 
Under the interval assumption, we take $T$ to be $[x^{L}, x^{U}]$ such that $\forall i \in [n], x^{L}_i \leq x_i \leq x^{U}_i$. For simplicity, we describe this interval with a width vector $\epsilon \in \mathbb{R}^{n}$ such that $[x^{L}, x^{U}] = [x - \epsilon, x + \epsilon]$. We now present the analogous definition for the model perturbation case.
\begin{definition}\label{def:modelcert} \textbf{\textit{Model-Robust Explanation}} Given a neural network $f^{\theta}$, a test input $x$, a loss function $\mathcal{L}$, a compact parameter set $M$, and a vector $\mathbf{\delta} \in \mathbb{R}^{n}$ we say that the explanation of the network $f$ is $\delta$-model-robust iff $\forall i \in [n]$: $\lVert \min_{\theta \in M} \partial \mathcal{L}/ \partial x_i - \max_{\theta \in M} \partial \mathcal{L}/ \partial x_i \rVert \leq \delta_i$. Moreover, we define $E_{M} := \{ v'\ |\ \forall i \in [n], \  \min_{\theta \in M} \partial \mathcal{L}/ \partial x_i \leq v'_i \leq  \max_{\theta \in M} \partial \mathcal{L} / \partial x_i \}$.
\end{definition}
This definition is identical to Definition~\ref{def:inputcert} save for the the fact that here the guarantee is over the model parameters. As such, a model that satisfies this definition guarantees that for the given neural network and input, there does not exist an adversary that can change the explanation outside of $E_{M}$ given that it only perturbs the parameters by amount specified by $M$. Similar to what is expressed for inputs, we express $M$ as an interval over parameter space parameterized by a width vector $\gamma \in \mathbb{R}^{n_p}$. To avoid issues of scale, we express an interval over a weight vector $W$ w.r.t. $\gamma$ as $[W - \gamma |W|, W + \gamma |W|]$. {\color{black} Before proceeding we remark on the choice of $\mathcal{L}$. Typically, one takes the input-gradient of the predicted class to be the explanation; however, one may want to generate the class-wise explanation for not just the predicted class, but other classes as well \citep{zhou2016learning, selvaraju2017grad}. In addition, adversarial robustness literature often considers attacks with respect to surrogate loss functions as valid attack scenarios \citep{carlini2017towards}. By keeping $\mathcal{L}$ flexible in our formulation, we model each of these possibilities.}

\subsection{From Robust Explanations to Certified Explanations}\label{subsec:certification_vector}

In this section, we describe the procedure for going from $\delta$-robustness (input or model robustness) to certification against targeted or untargeted attacks. For ease of notation we introduce the notation $v^{U}$ for the maximum gradient ($\max_{x \in T} \nabla_x \mathcal{L}(f^{\theta}(x))$ or $\max_{\theta \in M} \nabla_x \mathcal{L}(f^{\theta}(x))$) and $v^L$ for the minimum gradient ($\min_{x \in T} \nabla_x L(f^{\theta}(x))$ or $\min_{\theta \in M} \nabla_x \mathcal{L}(f^{\theta}(x))$). This notation leads to  $\delta = |v^{U} - v^{L}|$  and $E = [v^{L}, v^{U}]$ according to Definitions \ref{def:inputcert} and \ref{def:modelcert}. Below we discuss how to use these values to certify that no successful targeted or untargeted attack exists.

\textit{Certification of Robustness to Targeted Attacks:} To certify that no successful targeted attack exists it is sufficient to check that $\forall v' \in E, h(v', v^{\text{targ}}) > \tau$. In order to do so, we can define $v^{\text{cert}} := \argmin_{v \in E} h(v, v^{\text{targ}})$. 
If the inequality $h(v^{\text{cert}}, v^{\text{targ}}) > \tau$ holds, then we have proven that no successful adversary exists. Where $h$ is the mean squared error similarity, we can construct $v^{\text{cert}}$ from the set $E$ that maximizes the similarity by taking each component $v^{\text{cert}}_i$ to be exactly equal to $v^{\text{targ}}_i$ when $v^{\text{targ}}_i \in [v^{L}_i, v^{U}_i]$, we take $v^{\text{cert}}_i = v^U_i$ when $v^{\text{targ}}_i > v^{U}_i$ and we take $v^{\text{cert}}_i = v^{L}_i$ otherwise. This construction follows from the fact that minimizing the mean squared error between the two vectors is accomplished by minimizing the mean squared error for each dimension. 

\textit{Certification of Robustness to Untargeted Attacks:} Similar to the argument above, we can certify the robustness to untargeted attacks by showing that $\forall v' \in E, h(v, v') \leq \tau$ where $v$ is the original input gradient. As before, if the $v^{\text{cert}} := \argmax_{v' \in E} h(v, v')$ has mean squared error less than $\tau$ then we have certified that no successful untargeted attack exists. We construct $v^{\text{cert}}$ in this case by taking each component $v^{\text{cert}}_i = v^U_i$ when $|v^U_i - v_i| > |v_i - v^L_i|$ and $v^{\text{cert}}_i = v^L_i$ otherwise. 

\begin{figure}
    \centering
    \includegraphics[width=0.8\textwidth]{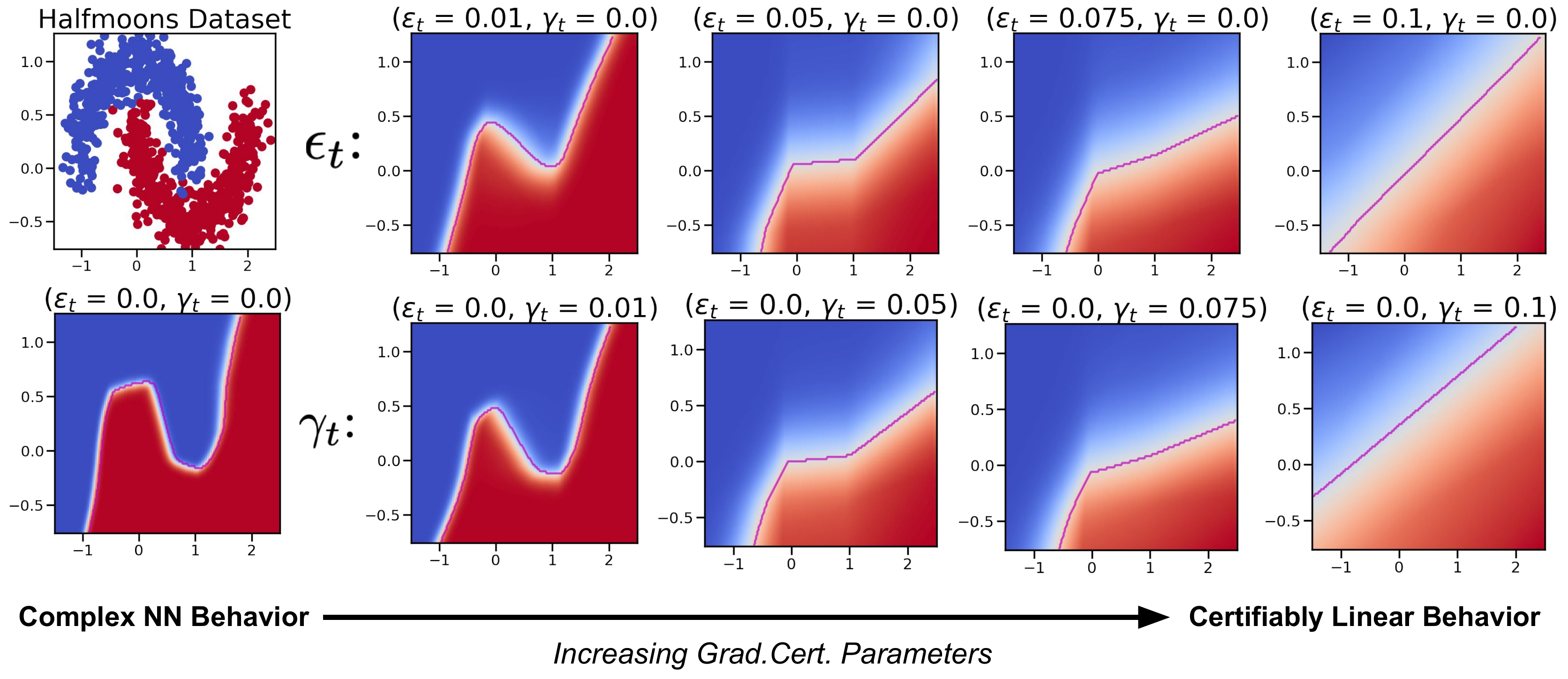}
    \caption{By increasing the parameters of our method, we are able to enforce certifiable locally-linear behavior of an NN model on the half-moons dataset, training data visualized in the top left corner. In the extreme case, illustrated on the far right, we recover a globally linear classifier. The top row varies $\epsilon_t$ the size of the input region while the bottom row varies $\gamma_t$ which enforces locally linear behavior w.r.t. the model parameters.
    }
    \label{fig:HalfmoonsBehavior}
\end{figure}

\section{Computational Framework}\label{sec:computations}

In this section we will provide a computationally efficient framework using interval bound propagation (IBP) to compute values for $v^L$ and $v^U$. We highlight that the values we compute are \textit{sound} but not \textit{complete}. In short, this means the values for $v^L$ computed here are not the largest values for the minimum that can be obtained and the values for $v^U$ are not the smallest values for the maximum that can be obtained. Tighter bounds may be achieved at the cost of greater computational complexity as discussed in Appendix~\ref{appendix:assumptions}; however the method we have chosen has computational cost equivalent to four forward passes through the network and is therefore very fast. 
In order to compute sound $v^{L}$ and $v^{U}$  we propagate input and/or model intervals through the forward and backwards pass of the neural network; however, standard propagation techniques do not accommodate joint propagation of input and weight intervals which is necessary for the backwards pass. To solve this problem, we present Lemma 1:  
%
\begin{lemma}\label{lemma:dualproplemma} \citep{wicker2021adversarial} Given an interval over matrices $A^{L}, A^{U} \in \mathbb{R}^{a \times b}$ s.t. $\forall i,j$ it holds that $A^{L}_{i,j} \leq A^{U}_{i,j}$, and an interval over matrices $B^{L}, B^{U} \in \mathbb{R}^{b \times c}$ s.t. $\forall i,j\ B^{L}_{i,j} \leq B^{U}_{i,j}$. We denote the center and width of a matrix interval as $B^{\mu} = (B^{L} + B^{U})/2$ and $B^{r} = (B^{U} - B^{L})/2$. We also denote $M^{B} = |A|B^{r}$ and correspondingly, $M^{A} = A^{r}|B|$.
We then have the following element-wise inequalities $\forall A^* \in [A^{L}, A^{U}]$ and $\forall B^* \in [B^{L}, B^{U}]$: $$\forall i,j \quad A^{\mu}B^{\mu}_{\  i,j} - M^{B}_{i,j} - M^{A}_{i,j} - Q_{i,j}  \leq  A^*B^*_{\ i,j} \leq  A^{\mu}B^{\mu}_{\  i,j} + M^{B}_{i,j} + M^{A}_{i,j} + Q_{i,j}$$
where {\color{black}$|\cdot|$ indicates element-wise absolute value} and $Q = |A^{r}| | B^{r} | $. We denote a function that returns these matrix multiplication upper and lower bounds given the four parameters above as: $\mathcal{L}(A^{L}, A^{U}, B^{L}, B^{U})$ for the lower bound and $\mathcal{U}(A^{L}, A^{U}, B^{L}, B^{U})$ for the upper bound.
\end{lemma}
As desired, Lemma~\ref{lemma:dualproplemma} allows us to jointly propagate intervals over inputs and parameters through affine transformations. We can now restate the forwards and backwards pass in terms of lower and upper bounds on input features and model parameters: 
\begin{equation*}\centering\footnotesize
\begin{aligned}[c]
    \text{Forward}& \text{ Pass w.r.t. Bounds:}\\
    \color{iceblue}{z^{(L,0)}}\ &\color{black}{= x^{L}} \quad  \color{firered}{z^{(U,0)}}\ \color{black}{= x^{U}}\\
    \color{iceblue}{\zeta^{(L, k+1)}} &=  \mathcal{L}(W^{(L,k)}, W^{(U,k)}, \color{iceblue}{z^{(L,k)}}, \color{firered}{z^{(U,k)}} \color{black}{)}  \\
    \color{firered}{\zeta^{(U, k+1)}} &=  \mathcal{U}(W^{(L,k)}, W^{(U,k)}, \color{iceblue}{z^{(L,k)}}, \color{firered}{z^{(U,k)}} \color{black}{)}  \\
    \color{iceblue}{z^{(L,k)}} &= \sigma(\color{iceblue}{\zeta^{(L,k)}} \color{black}{+ b^{(L,k)})} \\
    \color{firered}{z^{(U,k)}} &= \sigma(\color{firered}{\zeta^{(U,k)}} \color{black}{+ b^{(U,k)})} \\
\end{aligned}
\quad 
\begin{aligned}[c]
    \text{Backward}& \text{ Pass w.r.t. Bounds:}\\
    \color{iceblue}{d^{(L, K)}} &\color{black}{=} \sfrac{\partial L}{ \partial z^{(L,K)}} \quad   
    \color{firered}{d^{(U, K)}} \color{black}{=} \sfrac{\partial L}{ \partial z^{(U,K)}}  \\
    \color{iceblue}{\delta^{(L, k-1)}} &\color{black}{=}  \mathcal{L}\bigg(\dfrac{\partial z^{(L,k)}}{ \partial \zeta^{(k)}}, \dfrac{\partial z^{(U,k)}}{ \partial \zeta^{(k)}}, \dfrac{\partial \zeta^{(L,k)}}{ \partial z^{(k-1)}}, \dfrac{\partial \zeta^{(U,k)}}{ \partial z^{(k-1)}} \bigg)  \\
    \color{firered}{\delta^{(U, k-1)}} &\color{black}{=}  \mathcal{U}\bigg(\dfrac{\partial z^{(L,k)}}{ \partial \zeta^{(k)}}, \dfrac{\partial z^{(U,k)}}{ \partial \zeta^{(k)}}, \dfrac{\partial \zeta^{(L,k)}}{ \partial z^{(k-1)}}, \dfrac{\partial \zeta^{(U,k)}}{ \partial z^{(k-1)}} \bigg) \\
    \color{iceblue}{d^{(L,k-1)}} &\color{black}{=} \text{MaxMul}(\color{firered}{d^{(U, k)}}, \color{iceblue}{d^{(L, k)}}, \color{firered}{\delta^{(U, k-1)}}, \color{iceblue}{\delta^{(L, k-1)}} \color{black}{)}  \\
    \color{firered}{d^{(U,k-1)}} &\color{black}{=} \text{MinMul}(\color{firered}{d^{(U, k)}}, \color{iceblue}{d^{(L, k)}}, \color{firered}{\delta^{(U, k-1)}}, \color{iceblue}{\delta^{(L, k-1)}} \color{black}{)}  \\ 
\end{aligned}
\end{equation*}
%
where $\text{MaxMul}(A,B,C,D)$ is short-hand for the elementwise maximum of the products $\{ AB, AD, BC, BD\}$ and the $\text{MinMul}$ is defined analgously for the elementwise minimum. Following the above iterative equations we arrive at $d^{L, 0}$ (lower bound) and $d^{U, 0}$ (upper bound) which are upper and lower bounds on the input gradient for all inputs in $[x^{L}, x^{U}]$ and for all parameters in $[\theta^{L}, \theta^{U}]$. Formally, these bounds satisfy that  $\forall i \in [n], \ d^{L, 0}_i \leq v^{L}_i$ and $d^{U, 0}_i \geq v^{U}_i$. {\color{black} We provide more detailed exposition and proof in Appendix~\ref{appendix:proofs}}. Finally, we can write $\delta^{\text{cert}} = d^{U, 0} - d^{L, 0}$ and $E^{\text{cert}} = [d^{L, 0}, d^{U, 0}]$ which satisfy the conditions of Definition~\ref{def:inputcert} and \ref{def:modelcert}. 


\subsection{Gradient Certification Training Loss}
It follows from Definitions \ref{def:inputcert} and \ref{def:modelcert} that models with smaller $\delta$ values are more robust. To minimize $\delta$ during training, it is natural to add $\sum_{i=0}^{n-1} \delta_i$ as a regularization term to the network loss.
At training time, we consider the input perturbation budget $\epsilon_t$ as well as the model perturbation budget $\gamma_t$. The input interval is simply taken to be $[x-\epsilon_t, x+\epsilon_t]$ and the interval over a weight parameter $W$ is $[W - |W|\gamma_t, W + |W|\gamma_t]$. Given such intervals, we can then state a function $D(x, \theta, \epsilon_t, \gamma_t)$ which returns the vector $\sum_{i=0}^{n-1} \delta_i^{\text{cert}}$. Thus, minimizing $D(x, \theta, \epsilon_t, \gamma_t)$ corresponds to minimizing an upper bound on the maximum difference in the first derivative leading to improved certification. The complete, regularized loss which we term the gradient certification (Grad. Cert.) loss is:  
$$\mathcal{L}_{\text{Grad. Cert.}} = \mathcal{L}(f^{\theta}(x)) +  \alpha D(x, \theta, \epsilon_t, \gamma_t)$$
where $\alpha$ is a parameter that balances the standard loss and the adversarial loss. Throughout the paper, we will continue to refer to the values of $\epsilon$ and $\gamma$ that are used at training time as $\epsilon_t$ and $\gamma_t$.
The optimal loss for our regularization term is achieved for an input point $x$ when $D(x, \theta, \epsilon_t, \gamma_t) = 0$. For input robustness, implies that for all points in set $T$ the gradient with respect to the input is constant. Thus, one perspective on our regularization is that it enforces certifiably locally linear behavior of the neural network for all points in $T$. For $\gamma_t$ it is a similar guarantee save that all of the models in $M$ must have point-wise linear behavior at $x$. 
In Figure~\ref{fig:HalfmoonsBehavior} we visualize the effect of taking our regularization term to the extreme in a planar classification problem. We see that indeed this perspective holds and that increasing the parameters $\epsilon_t$ or $\gamma_t$ leads to the classifier having more linear behavior.

\section{Experiments}
In this section, we empirically validate our approach to certification of explanation robustness as well as our regularization approach to training NNs with certifiably robust explanations. We start by covering the evaluation metrics we will use throughout this section to benchmark robustness of explanations. We analyze robustness of explanations on two financial benchmarks, the MNIST handwritten digit dataset, and two medical image datasets. We report exact training hyper-parameters and details in the Appendix along with further ablation studies, Appendix~\ref{appendix:regularization-comparison}, \ref{appendix:robustness}, \ref{appendix:fairness}, \ref{appendix:medmnist}, \ref{appendix:extendedcomparison}, {\color{black} and discussion of assumptions and limitations in Appendix~\ref{appendix:assumptions}}. 

\textbf{Evaluation Metrics} 
In this section we state the metrics that we will use throughout our analysis. Each of the metrics below are stated for a set of $j$ test inputs $\{x^{(i)}\}_{i=0}^{j-1}$ and for each application we will explicitly state the $\tau$ used.

\textit{Attack robustness:} Given a neural network $f^{\theta}$, we denote $v^{(i),\text{adv}}$ as the input gradient resulting from an attack method on the test input $x^{(i)}$. We then define the metric as: $1/j \sum_{i=0}^{j-1} \mathds{1}(h(v^{\text{targ}}, v^{(i), \text{adv}}) > \tau)$. When $v^{(i), \text{adv}}$ is the result of \citep{dombrowski2019explanations}, a first-order attack in the input space using PGD, we call this \textit{input attack robustness}. Whereas if $v^{(i), \text{adv}}$ is the result of a locally bounded version of the attack in \citep{heo2019fooling}, a first-order attack in the parameter space using PGD, we call this \textit{model attack robustness}. Intuitively, this measures the proportion of inputs that are robust to various attack methods.

\textit{Certified robustness:} Using our certification method, we compute $v^{(i), \text{cert}}$ according to Section~\ref{subsec:certification_vector} and then measure: $1/j \sum_{i=0}^{j-1} \mathds{1}(h(v^{\text{targ}}, v^{(i),\text{cert}}_i) > \tau)$. Intuitively, this metric is the proportion of inputs which are certified to be robust. Where $v^{(i), \text{cert}}$ is the result of propagating an input set $T$ this is \textit{input certified robustness}, and when propagating $M$ we call it \textit{model certified robustness}.

\begin{figure}
    \centering
    \includegraphics[width=0.8\textwidth]{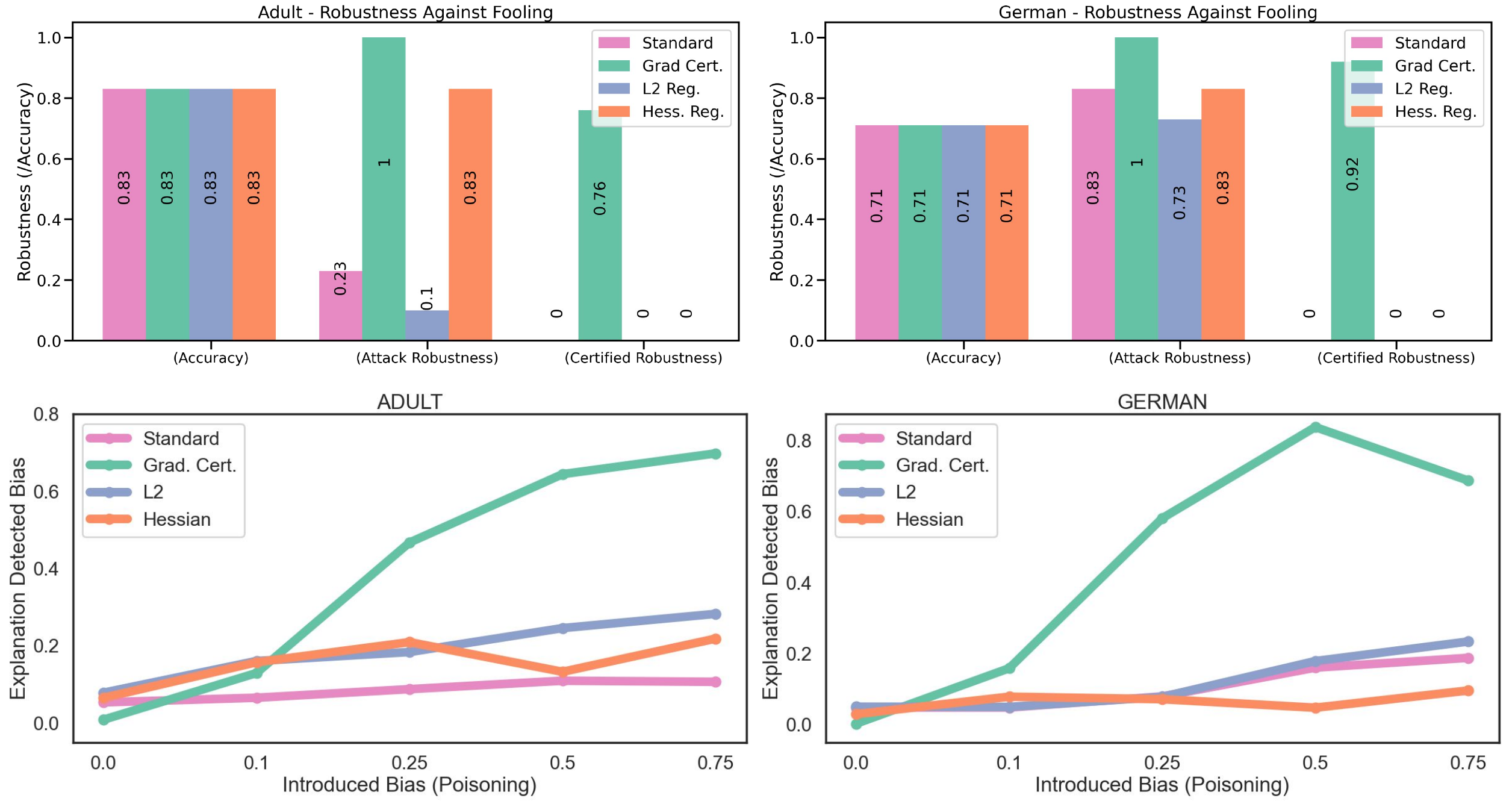}
    \caption{\textbf{Top Row:} Only training with our gradient certified loss gives non-trivial certified robustness performance (bars labelled \textit{Certified Robustness}). We find no drop in accuracy when using our method (bars labelled \textit{Accuracy}), and in addition we have attack robustness that surpasses baseline models (bars labelled \textit{Attack Robustness}). \textbf{Bottom Row:} We introduce a variable amount of label poisoning (x-axis) to introduce bias into the datasets, and then test to see if the explanation indicates that it is using sensitive features. Looking at the magnitude of the gradients, we observe that gradient certified training forces the model to reveal its sensitivity much more readily. } 
    \label{fig:financeresults}
\end{figure}

\subsection{Certifying Explanations in Financial Datasets}

In financial applications, it is of interest that users not be able to game financial models into providing incorrect explanations by slightly changing their features. We study the robustness of NNs in financial applications to adversarial users attempting to game the NN into providing an explanation that shows the model is predicting on the basis of protected attributes (e.g., gender, race, age). In order to do so we study the \textit{Adult} dataset, predicting whether or not an individual makes more or less than fifty thousand dollars a year, and the \textit{German Credit} dataset, predicting whether an individual has a good or bad credit rating. Importantly, financial modellers should rigorously ensure/certify that their models are indeed not biased \citep{benussi2022individual}; we discuss this further in Appendix~\ref{appendix:fairness}. We assess the robustness against gaming by taking $v^{\text{targ}}$ to be a vector with one entries for each of the sensitive features and zeros elsewhere, and we say an attack is successful if the adversary is able to force the sensitive feature (gender) into being in the top 5 explaining features by magnitude. That is, an attack is successful if the explanation reveals that the sensitive feature is one of the most relied upon features. 
In the top row of Figure~\ref{fig:financeresults} we present the accuracy and robustness values for standard training, gradient certified training (ours), and other robust explanation baselines including those suggested in \citep{dombrowski2022towards}, Hessian regularization + softplus activation functions, and \citep{drucker1992improving}, training to minimize $\ell_2$ gradient difference in a ball around the inputs. {\color{black} Objectives, further baselines, and details can be found in Appendix~\ref{appendix:extendedcomparison}.} On the Adult dataset we find that networks trained with \citep{drucker1992improving} regularization and without any regularization (standard) are vulnerable to explanation attacks. On the German dataset, we find that overall, models are more robust to being fooled. However, we consistently find that the best robustness performance and the only certified behavior comes from gradient certified training. 

\textit{Bias detection study} We study the ability of explanations to flag bias in models. We create biased models by performing label poisoning. We set a random proportion $p$ of individuals from the majority class to have a positive classification and the same proportion $p$ of individuals from the minority class to have a negative classification. The proportion $p$ is labeled `induced bias' in Figure~\ref{fig:financeresults} is strongly correlated with the actual bias of the model, see Appendix~\ref{appendix:fairness}. 
We then use the input gradient, $v$, to indicate if the explanation detects the bias in the learned model. This is done by measuring the proportion $|v_j|/\sum_{i=0}^{n-1} |v_i|$ where $j$ is the index of the sensitive feature this proportion is called `explanation detected bias' in our plots. We highlight that our method significantly out-performs other methods at detecting bias in the model likely due to the certified local linearity providing more calibrated explanations. While our method is promising for bias detection it should supplement rather than replace a thorough fairness evaluation \citep{barocas2016big}. 

\begin{figure}
    \centering
    \includegraphics[width=0.85\textwidth]{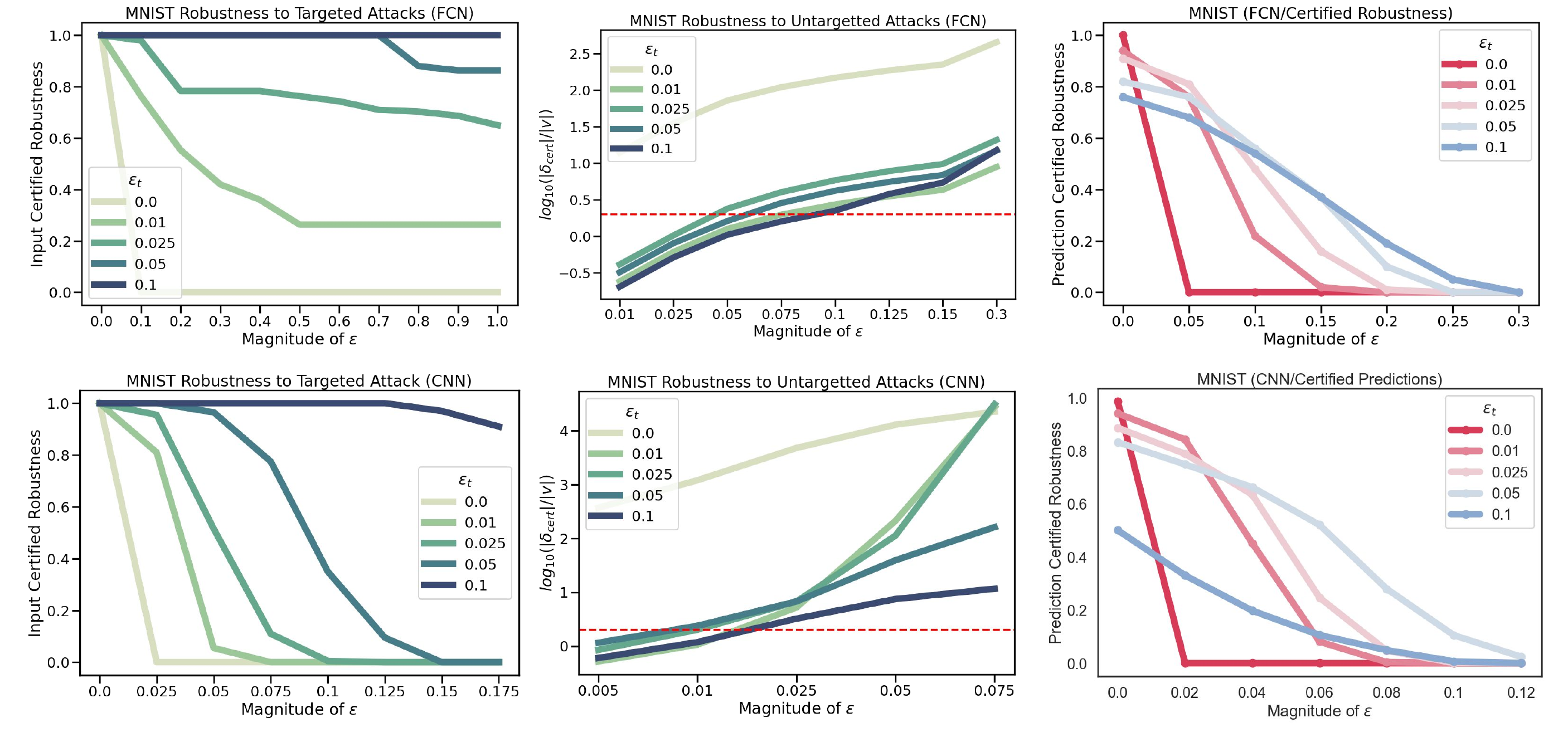}
    \caption{Gradient certified training leads to neural networks with considerably heightened robustness against targeted (left column) and untargeted attacks (center column). In addition, we find that our training leads to certifiably robust predictions (right column). \textbf{Left column: } Targeted attack robustness, higher is better, against adversary who tries to force explanation into the corner of an image. We see that FCNs are robust to this attack over the entire domain ($\epsilon = 1.0$). \textbf{Center column: } On the y-axis we plot the the average value of $\delta$ normalized by the gradient magnitude, lower is better. We plot a red dashed line at the success threshold for untargeted attacks and find that FCNs are considerably more robust to untargeted attacks. \textbf{Right column: } On the y-axis we plot the certified robust accuracy, showing that gradient certified training leads to certified prediction robustness. }
    \label{fig:mnistresults}
\end{figure}

\subsection{Certifying Explanations in Digit Recognition} 
In this section, we study the affect of input peturbations on the explanations for networks trained on the MNIST digit recognition dataset. 
We train two networks, a two-layer fully-connected neural network with 128 neurons per layer and a four layer convolutional neural network inspired by the architecture in \citep{gowal2018effectiveness}.
We take a set of twenty target explanations to be binary masks with a five by five grid of ones in the corners with varying distances from the border of the image and based on visual inspection set $\tau = 0.04$, see Figure~\ref{fig:ThreatModels} in the Appendix for illustration. In Figure~\ref{fig:mnistresults}, we plot the result of our robustness analysis. We note that there is a consistent $2\%$ test-set accuracy penalty for every $0.01$ increase in $\epsilon_t$. We find that for both FCNs and CNNs we are able to get strong certification performance. We highlight that for FCNs we are able to certify that there is no input in the entire domain that puts its explanation in any corner of the input image, this is indicated by $\epsilon = 1.0$, as the images are normalized in the range $[0,1]$. 
In the center column of Figure~\ref{fig:mnistresults}, we plot the results of our untargeted attack analysis. In particular we take the untargeted $v^{\text{cert}}$ described in Section~\ref{subsec:certification_vector} and measure $||v^{\text{cert}}||_2/||v||_2$. 
Again based on visual inspection, we set the untargeted threshold $\tau = 2.0$, indicated with a red dashed line. 
Finally, in the right-most column of Figure~\ref{fig:mnistresults}, we plot how our training method affects the certifiable prediction robustness.
Using bounds from \citep{gowal2018effectiveness} we can certify that no input causes the prediction to change. We show that for both FCNs and CNNs our bounds lead to considerable, non-trivial certified prediction robustness.

\begin{figure}
    \centering
    \includegraphics[width=0.75\textwidth]{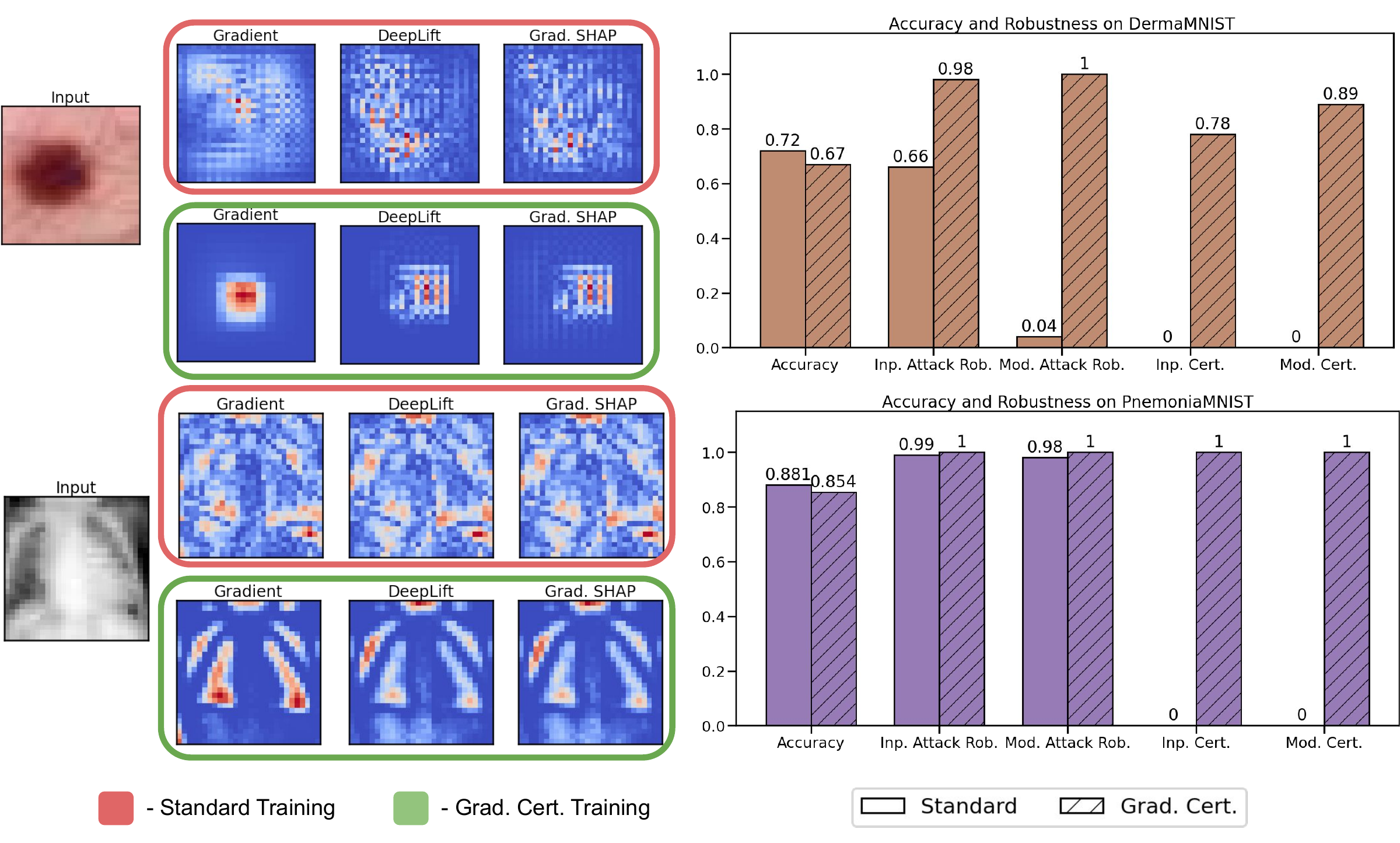}
    \caption{On MedMNIST datasets the gradient certified loss improves attack robustness and gets strong input and model certified robustness. In addition, we find that explanations from a variety of methods are more sparse and intuitive than explanations from normally trained networks. \textbf{Top rows: } On DermaMNIST, the gradient certified loss leads to considerable increases in all robustness metrics for only a 5\% reduction in test-set accuracy. \textbf{Bottom rows: } On PneumoniaMNIST despite standard training having strong attack robustness performance only gradient certified training has non-trival certified robustness with only a 3\% reduction in test-set accuracy.   
    }
    \label{fig:medmnistresults}
\end{figure}

\subsection{Certifying Explanations in Medical Image Classification}

We consider two datasets from the MedMNIST benchmark and a third in Appendix~\ref{appendix:medmnist}. The first task, \textit{DermaMNIST}, is to classify images into one of 11 different skin conditions. The second task, \textit{PneumoniaMNIST}, is to classify chest x-rays into positive or negative diagnosis for pneumonia. DermaMNIST is represented by full-color images (2352 feature dimensions) while PneumoniaMNIST is represented by black and white images (784 feature dimensions). We use the same 20 binary corner masks as in the MNIST analysis. We train models with $\epsilon_t = 0.01$ and $\gamma_t = 0.01$ and test with $\epsilon = 0.025$ and $\gamma = 0.025$. Across each dataset, we find that the only non-trivial certification comes from training with our method. Even for PneumoniaMNIST where the standard network shows impressive attack robustness, we are unable to certify explanation robustness for standard trained models. 
This result mirrors the fact that only certified training yields non-vacuous prediction robustness bounds for high-dimensional inputs in the adversarial attack literature \citep{gowal2018effectiveness}. We highlight that our method indicates that no model attack can be successful if it perturbs the weights by $2.5\%$ of their value, but we cannot make guarantees outside of that range, and in principle there will always exist a classifier outside of the guaranteed range that corrupts the input gradient. In the left half of Figure~\ref{fig:medmnistresults}, we plot example explanations from NNs trained with standard training as well as our training method. To avoid cherry-picking bias we use the first test image for each dataset. 
We find that the explanations provided by networks trained with our method are more sparse and seem to be better correlated with intuitively important features of the input image.

\section{Conclusion}

We present a computational framework for upper-bounding the largest change an adversary can make to the input gradient of a neural network given bounded modification of the input features or model parameters. Our framework allows us to present, to our knowledge, the first formal certification of explanation robustness for general neural networks. We empirically validate our findings over six datasets of varying complexity and find that our method is able to produce certificates of explanation robustness for each.

\section*{Acknowledgments}
MW acknowledges support from Accenture. MW and AW acknowledge support from EPSRC grant EP/V056883/1. 
AW acknowledges support from a Turing AI Fellowship under grant EP/V025279/1, and the Leverhulme Trust via CFI.

\bibliography{iclr2021_conference}
\bibliographystyle{iclr2021_conference}
\appendix
\clearpage
\newpage
\appendix

\section{Discussion of Methodological Assumptions}\label{appendix:assumptions}
\begin{figure}
    \centering
    \includegraphics[width=1.0\textwidth]{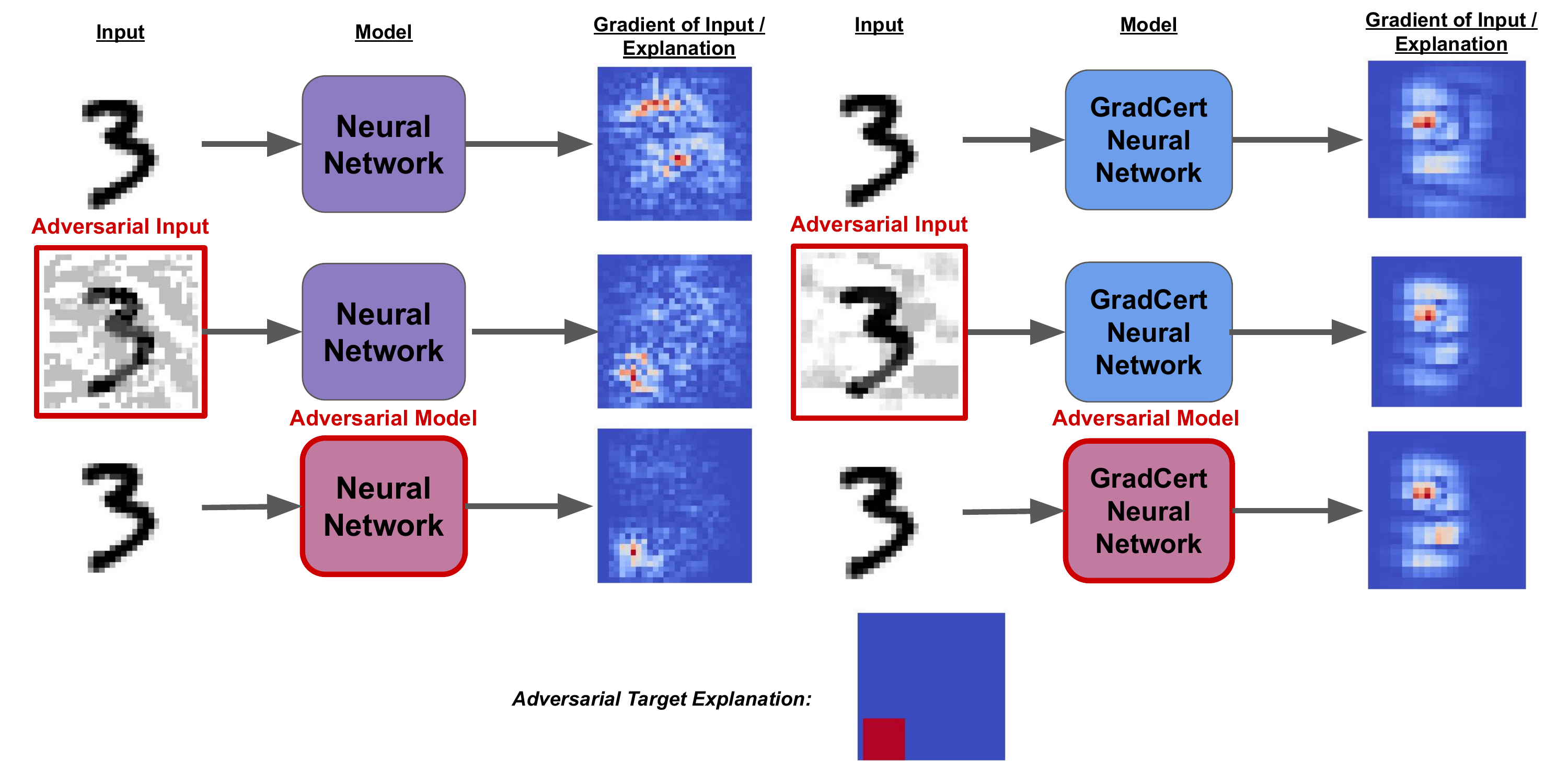}
    \caption{Example of threat models for explanation methods and their affects on a normally trained network versus a network trained with the gradient certification loss. \textbf{Top row}: the explanations given for an input classified as a `3' for a normally trained neural network (left) and for the GradCert network (right). \textbf{Center row}: The result of using the first-order attack proposed in \cite{dombrowski2019explanations} on the normal neural network (left) where it is successful and the GradCert (right) where it is unsuccessful. \textbf{Bottom row}: The result of using a single-image version of \cite{heo2019fooling} on a normal neural network (left) and a GradCert neural network (right). }
    \label{fig:ThreatModels}
\end{figure}

In the main text of the paper our primary methodological assumptions included assumptions on the monotonicity of activation functions, architecture of the neural network, and representation of the input and parameter sets. Below, we discuss how and when these assumptions can be relaxed. We also visualize the various threat models and attack targets in Figure~\ref{fig:ThreatModels}.

\paragraph{Activation Monotonicity} Activation functions being monotonic is important to the method as input extrema are also output extrema. That is, if the function is monotonically increasing (as is the case with ReLu, Tanh, Sigmoid) then evaluating the activation function at the maximum and minimum values for the domain will correspond to the maximum and minimum values for the range. When an activation is monotonically decreasing, the maximum of the domain interval will correspond to the minimum of the range interval and visa versa. While this makes exposition of the method convenient we note that our propagation method can be extended to activation functions with a bounded number of local maxima and minima. By breaking the function up at these points we get a series of monotonically increasing and decreasing segments and computing the extrema over all such segments suffices for computing the extrema over the range of the function.

\paragraph{Architectural Assumptions} In the main text we assume that all layers of the neural network are affine transformations followed by the application of a non-linear activation function. This is generally true of sound certification procedures (certified smoothing not being considered wholly sound, but statistically sound). Our assumption covers convolutional and fully connected layers, but does not handle pooling and batch-norm layers as these layers in their natural formulation are not amenable to reverse propagation with bound propagation methods. This remains an area of study that can enhance bound propagation methods. {\color{black} We are unable to scale to large neural network architectures such as VGG16 due to the fact that current methods for convex relaxation of such complex networks incur too much approximation for non-trivial certification \cite{gowal2018effectiveness}. As convex relaxation for neural networks advance our presented methodology for certifying explanation robustness will also advance.}  

\paragraph{Input and Parameter Set Representation} In our exposition we only provide a formulation for intervals over inputs and parameter sets. Along with \citep{gowal2018effectiveness}, we find that the interval representation of constraints is sufficient. However, there are more complex input properties which may not be captured well by this assumption, and there is research on more expressive abstractions. For a full treatment we reference interested readers to \citep{gehr2018ai2}. In \cite{gehr2018ai2} the authors discuss and evalute different approximations for verifying properties of neural networks including the box domain (called intervals in this work), zonotope domain, and the polyhedra domain. In general, more complex abstract domains allow expression of more complex properties.

\paragraph{Bound Propagation Technique} In Section~\ref{sec:computations} we point out that our bounds are not the tightest bounds that can be achieved for the interval domain. There have been many advances in bound propagation for sound certification of neural networks \cite{tjeng2017evaluating, gehr2018ai2, fazlyab2019safety, benussi2022individual}. More complex propagation techniques arrive at tighter bounds, in our case on the values of $\delta$ and $E$. For example MILP formulations are exact in the limit of computational time and refinement iterations \cite{benussi2022individual}. However, it is well known that MILP is exponentially more expensive than the methods presented in this paper. The specifications and techniques in this work can be adapted to any solver, and some solvers give tighter bounds at the cost of increased computational complexity. We have chosen IBP for this method as its computational efficiency allows us to scale to CNNs with over half a million parameters which is infeasable with MILP methods \citep{benussi2022individual}.

\begin{figure}
    \centering
    \includegraphics[width=0.8\textwidth]{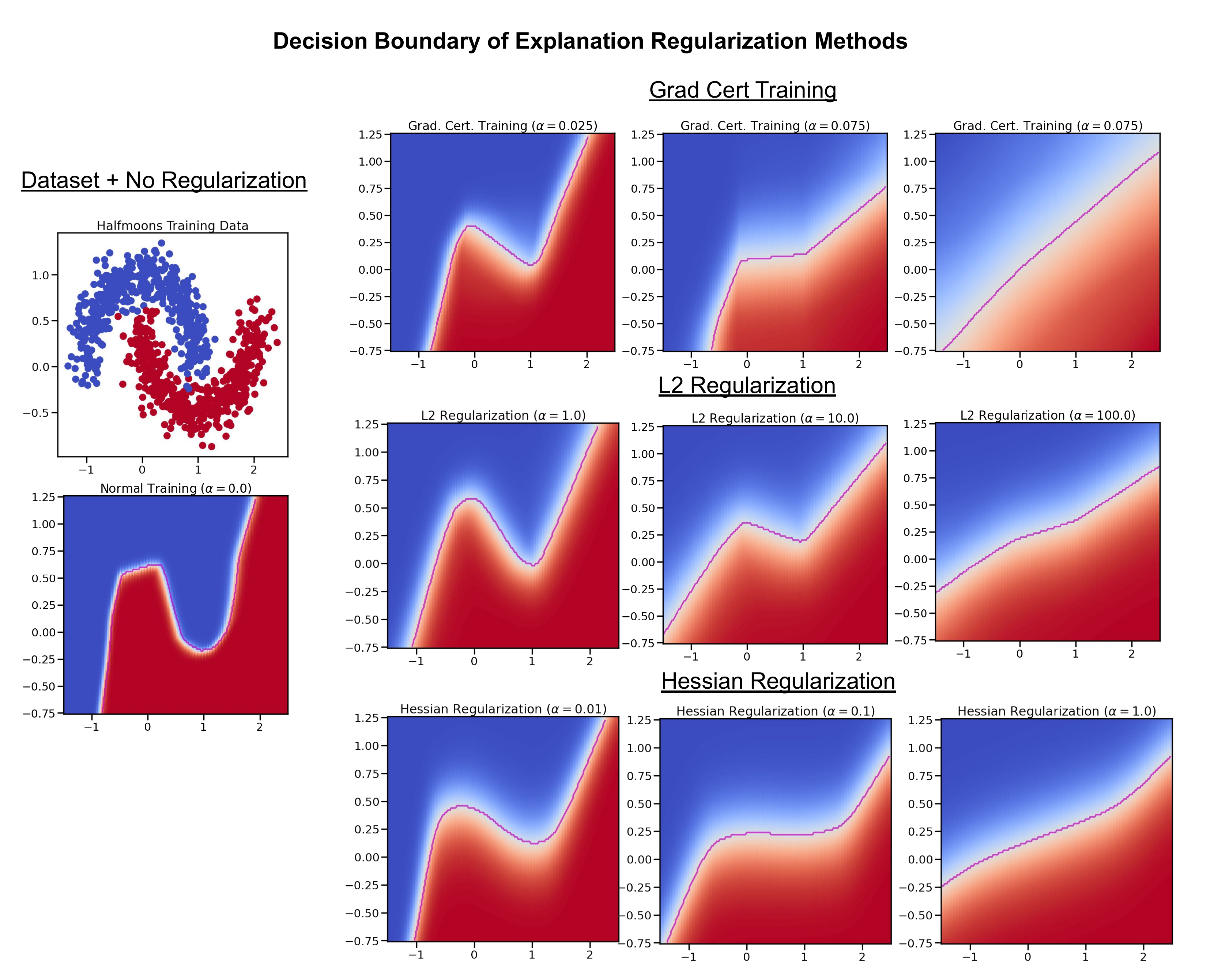}
    \caption{An empirical comparison of how different robust explanation regularization affect the decision boundaries of learned neural networks. From right to left we increase the regularization parameter (denoted $\alpha$ in the main text). \textbf{Top Row:} Grad. cert. regularization. \textbf{Middle Row:} L2 regularization \cite{chen2019robust}. \textbf{Bottom Row:} Hessian regularization \cite{dombrowski2019explanations}.}
    \label{fig:halfmoons-compare}
\end{figure}

\section{Effect of Explanation Regularization Methods}\label{appendix:regularization-comparison}

In this section we perform an empirical study of the effects of robust explanations training methods, visualized in Figure~\ref{fig:halfmoons-compare}. For this, we revisit the half-moons dataset and we train classifiers with our method, Hessian regularization \citep{dombrowski2019explanations}, and $L2$ adversarial training \citep{chen2019robust}. We find that each of the methods smooth the clssification boundary. We highlight that while our method and that of \citep{chen2019robust} enforce local linearity, Hessian regularization enforces smoothness but not linearity. We notice that our method is slightly more linear than the other methods for extreme values of the regularization parameters, but in practice, such regularization would not occur due to the considerable performance trade-off that comes with using a linear model. However, we do notice that for more complex datasets the models do have different robustness performance and we find that only our method is able to train NNs with certifiable explanation robustness, see Figure~\ref{fig:financeresults}.

\begin{figure}
    \centering
    \includegraphics[width=0.75\textwidth]{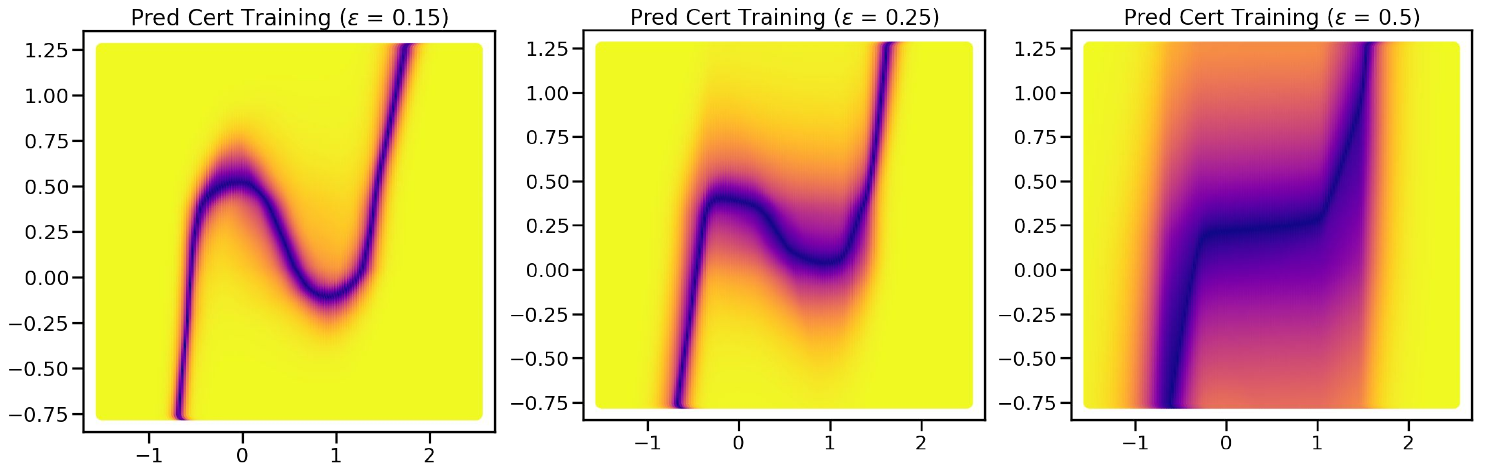}
    \caption{Decision boundaries and confidences for neural networks trained with certified prediction training using \citep{gowal2018effectiveness}. We highlight that this method does not have the linearizing effect observed for robust explanations methods and may even lead to misleading explanations when relying on the input gradients as we demonstrate the box in the far right plot.}
    \label{fig:halfmoons-robust}
\end{figure}

\begin{figure}
    \centering
    \includegraphics[width=1.0\textwidth]{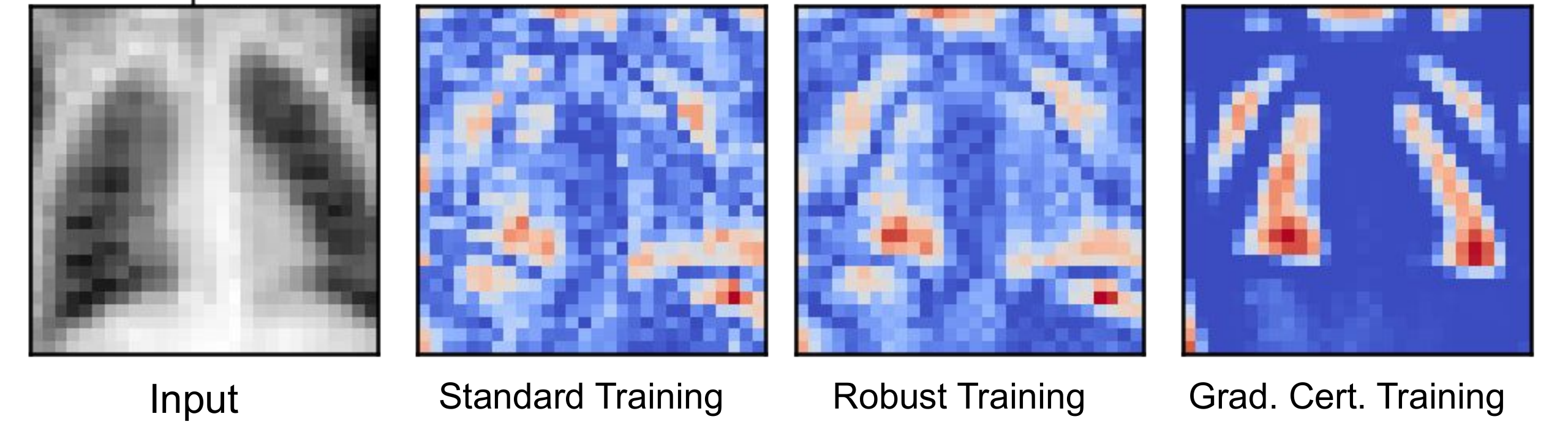}
    \caption{When visualizing the input gradients for different training methods, we see that the prediction robustness training using \citep{gowal2018effectiveness} leads to more sparse explanations (center gradient), but is not nearly as effective as gradient certified training (far right gradient). }
    \label{fig:appendix-pneumonia}
\end{figure}

\section{Effect of Robust Predictions Training}\label{appendix:robustness}

In this section we empirically compare our training with prediction robustness training. Robust prediction training \citep{gowal2018effectiveness, mirman2018differentiable} ensures that the prediction does not change for any point in the input interval whereas our method ensures the gradient doesn't change inside of the input interval and is unconstrained by the ground truth input. We compare our method with certified robust prediction training proposed in \citep{gowal2018effectiveness}. In Figure~\ref{fig:halfmoons-robust} we plot the resulting neural network decision boundary for different values of $\alpha$ when training with certified prediction robustness. We highlight that for large values of the regularization parameter $\alpha$ the robust prediction method does not induce a linear classifier. We also place a grey box on the far right plot to indicate where the network explanation based on the input gradient may be misleading. We also run the \citep{gowal2018effectiveness} training methodology on PneumoniaMNIST. We plot the resulting input gradients in Figure~\ref{fig:appendix-pneumonia}. We find that certified robust training does lead to sparser input gradients compared with standard training, but our training procedure is noticeably sparser. We also evaluate the average value of $\delta_i$ over 100 test-set images for each method and find that standard training has an average $\delta = 10.199$, robust prediction training has an average $\delta = 2.035$, and our method has an average $\delta = 0.026$. Thus, while robust training makes explanations considerably more robust, there is still a two orders of magnitude gap between robust explanations training and robust predictions training.

\begin{figure}
    \centering
    \includegraphics[width=0.75\textwidth]{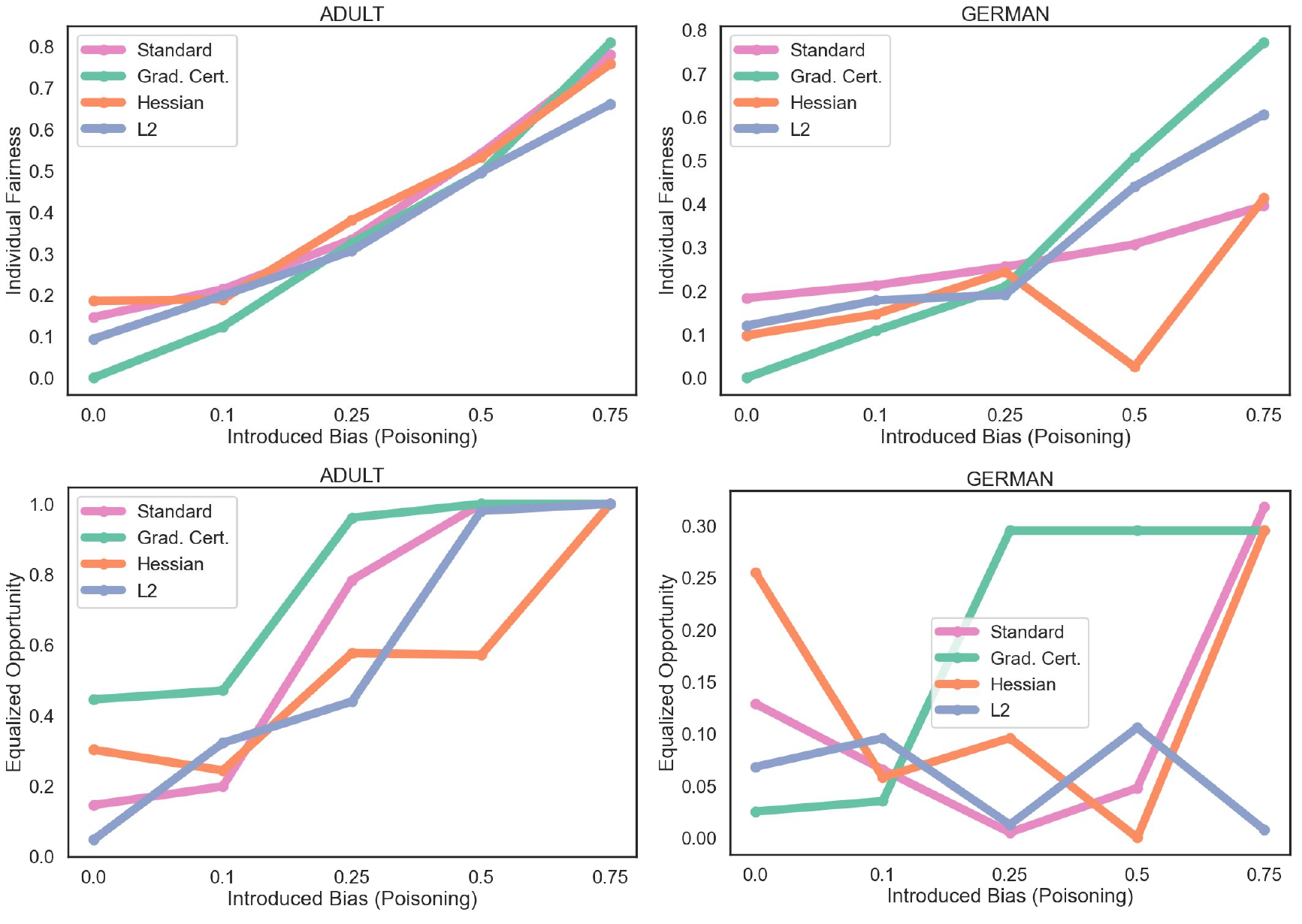}
    \caption{We see that our label poisoning has the intended effect of increasing the bias/unfairness of the classifier. \textbf{Top row: } We plot the individual fairness of the classifiers as we increase the poisoning rate. Higher individual fairness values indicate more bias. \textbf{Bottom row: } We plot the equalized opportunity, a group fairness metric, and find that for this metric we also increase bias for the Adult dataset, but it is less clear for the German dataset.}
    \label{fig:fairness-analysis}
\end{figure}

\section{Further Fairness Discussion}\label{appendix:fairness}

In order to detect bias, we first train classifiers on dataset in which we introduce bias. In order to introduce bias we take a random proportion $p \in [0,1]$ of individuals from the majority and minority classes and poison their labels, we call the proportion $p$ the `induced bias'. For the proportion $p$ of individuals in the majority class we set the label to a positive classification and for proportion $p$ of individuals in the minority class we change the label to a negative classification. The key idea here is that the neural network will pick up on the correlation between the majority/minority features and the label and will predict based on the sensitive attribute. In Figure~\ref{fig:fairness-analysis} we plot measures of individual fairness \cite{benussi2022individual} and group fairness \cite{binns2020apparent}. Individual fairness measures the difference in output for individuals who are  comparable, in our case, only differ by values of their sensitive output. For individual fairness what is plotted is the difference in softmax classification vectors for individuals (taken from the test set) who are identical save for their sensitive features. Higher values indicate more unfairness. We highlight that increases in discrimination according to individual fairness metrics rise with our increase in induced bias. In the bottom row of  Figure~\ref{fig:fairness-analysis} we plot a notion of group fairness which measures statistically how similar different majority and minority groups are treated, again higher values indicate more discrimination. We find that for the Adult dataset, group fairness metrics increase with increases in `induced bias' but for the German dataset it is less clear of a trend.

We would like to urge all readers that algorithmic fairness is a serious issue. While our method for robust explainability training seems to help indicate when the model is making bias predictions and may supplement fairness analysis, it should not substitute a rigorous fairness evaluation at train time and continuous fairness audits.

\begin{figure}
    \centering
    \includegraphics[width=0.95\textwidth]{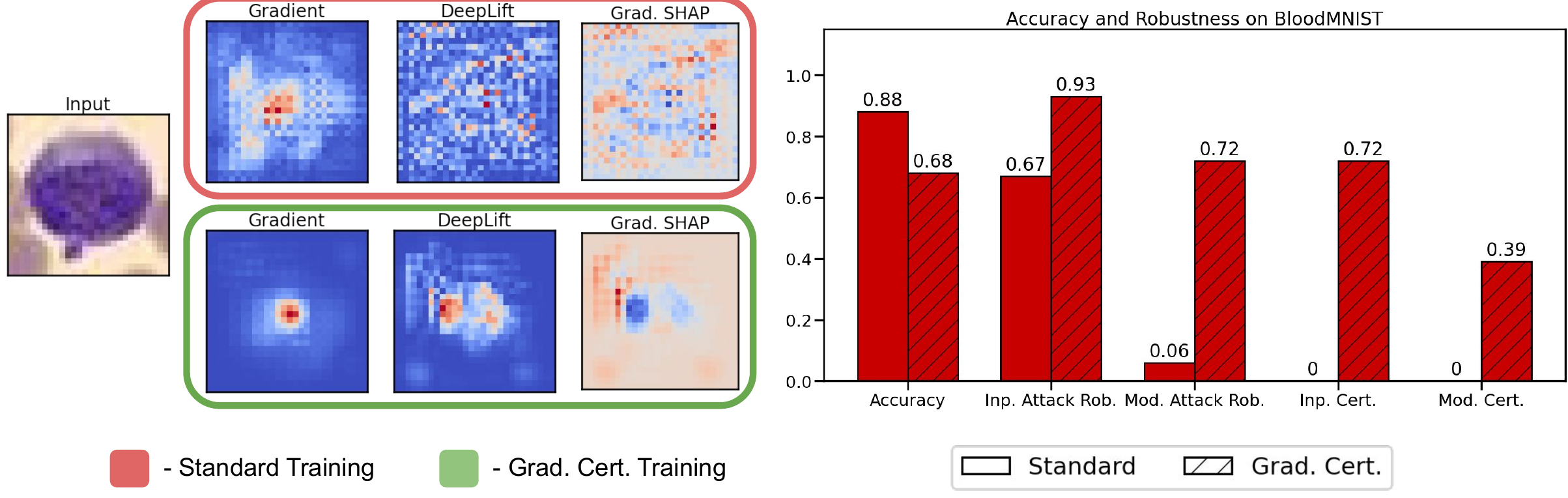}
    \caption{Analysis of three MedMNIST datasets. \textbf{Top Row:} Explanations on a test set DermaMNIST image, Grad Cert explanations (outlined in green) are nicely correlated with the key input features. The robustness and accuracy statistics indicate that there is a considerable 20\% reduction in test-set accuracy we gain substantial robustness benefits (from 0\% to 73\% input certified robustness). This figure extends Figure~\ref{fig:medmnistresults} in the main text.}
    \label{fig:appendix-medmnist}
\end{figure}

\section{Extended MedMNIST Results}\label{appendix:medmnist}

We extend the evaluation of our method on larger scale datasets by exploreing the BloodMNIST dataset which consists of 28 by 28 full-color images of stained blood cells representing 8 different blood disorders. The task of the neural network is to classify these disorders. In Figure~\ref{fig:appendix-medmnist} we find that we are able to significantly improve both the sparsity of the explanations as well as the well as the robustness of the network; however, it comes at the largest test set accuracy penalty of any dataset tested at 20\% test set accuracy loss. We note that further study into hyper-parameter tuning could significantly improve this result.

\section{Hyper-parameter Values}\label{appendix:hyperparams}

In this section we report the hyperparameters for the networks trained in the main text. Code to reproduce all of the experiments can be found at \textit{https://github.com/matthewwicker/RobustExplanationConstraintsForNeuralNetworks}. 

\subsection{Tabular Hyperparameters} For each dataset we use a fully-connected neural network with two layers and 256 hidden neurons per layer. Each network uses ReLU activation functions save for the Hessian training which uses softplus activations. Both our method and the L2 method \citep{chen2019robust} use $\alpha = 1.0$ and Hessian regularization \citep{dombrowski2019explanations} uses $\alpha = 0.1$ as using larger values leads to significant performance drops. We see that in Figure~\ref{fig:halfmoons-compare} that Hessian training is more sensitive to large changes in $\alpha$. Each network is trained for 100 epochs.

\subsection{MNIST Hyperparameters} We split the MNIST hyperparameter section into sections for the fully connected and convolutional networks. As we use the same architecture for the convolutional networks on MedMNIST we report the CNN hyperparameters in the MedMNIST section.

\textit{Fully Connected Networks: } We use neural networks with two hidden layers and 256 hidden neurons per layer. As before, we use ReLu activation functions. We set $\alpha = 0.5$ for each and vary $\epsilon_t$ as shown in Figure~\ref{fig:mnistresults}. Each network is trained for 35 epochs.

\subsection{MEDMNIST Hyperparameters} Below we describe the training parameters used for our MedMNIST experiments.

\textit{Convolutional Neural Networks: } We consider a small CNN model that was proposed in \citep{gowal2018effectiveness}. The network contains of two convolutional layers with 4 by 4 filters. The first layer consists of 16 filters and the second consists of 32 filters. We then flatten the features and pass it to a fully connected hidden layer with 100 neurons. Each layer uses ReLu actiavtions. For MNIST we set $\alpha = 0.5$ for each and vary $\epsilon_t$ as shown in Figure~\ref{fig:mnistresults}. Each network is trained for 35 epochs. For MedMNIST we keep $\alpha = 0.5$ and set $\epsilon_t = 0.01$ and $\gamma_t = 0.01$. We found empirically that these gave good results without dropping accuracy.

\begin{figure}[H]
    \centering
    \includegraphics[width=0.9\textwidth]{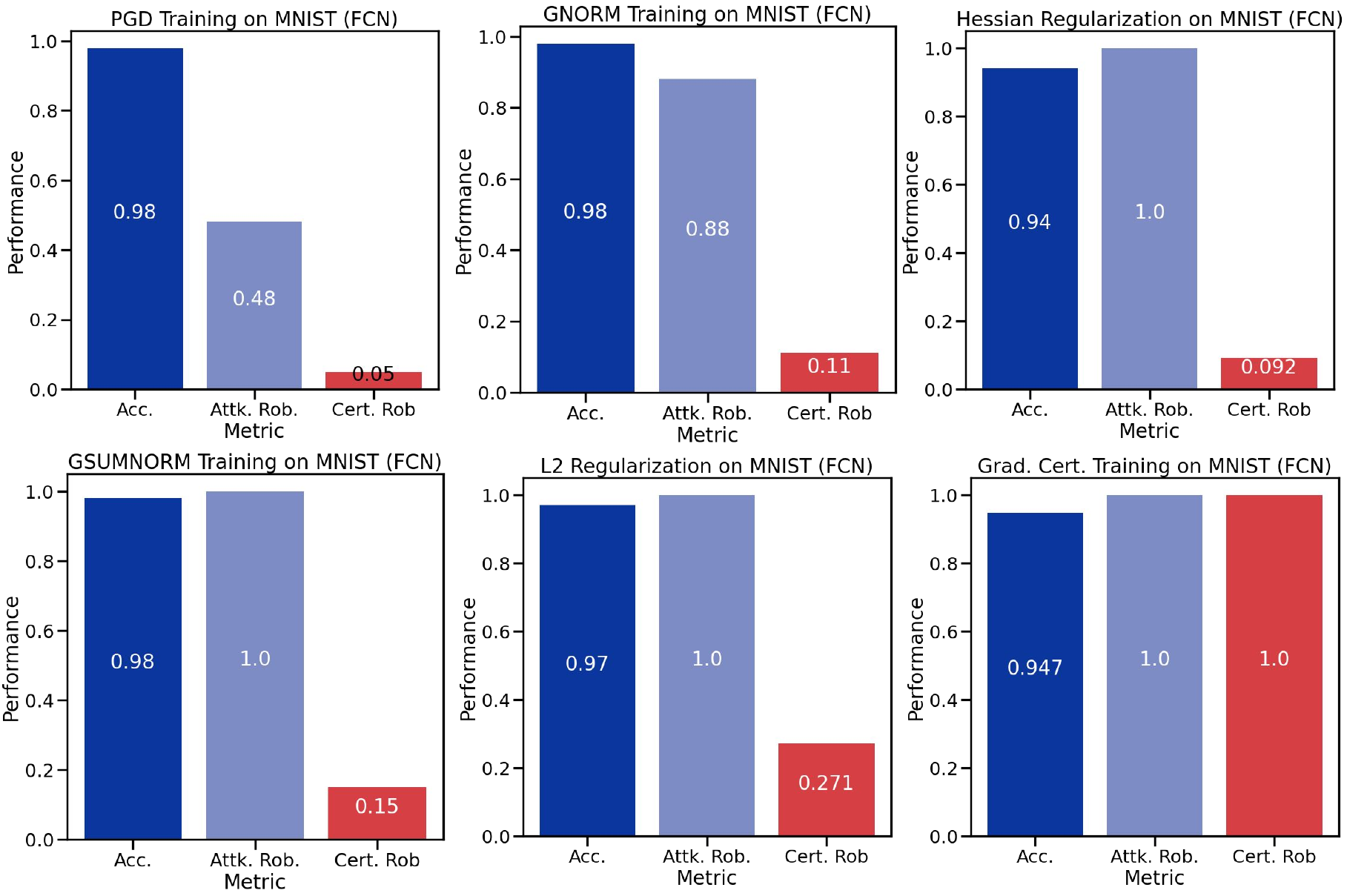}
    \caption{Our method out performs other robust explanation regularization approaches on the MNIST dataset. On the far left we plot standard training, center left we plot Hessian regularization, center right we plot L2 regularization, and far right we plot our method. We find that though regularization methods improve robustness against attacks and even provide some non-trivial certification in the case of L2 regularization, our method considerably out-performs each method with limited accuracy penalty.}
    \label{fig:mnist-compare}
\end{figure}

\begin{figure}[H]
    \centering
    \includegraphics[width=0.9\textwidth]{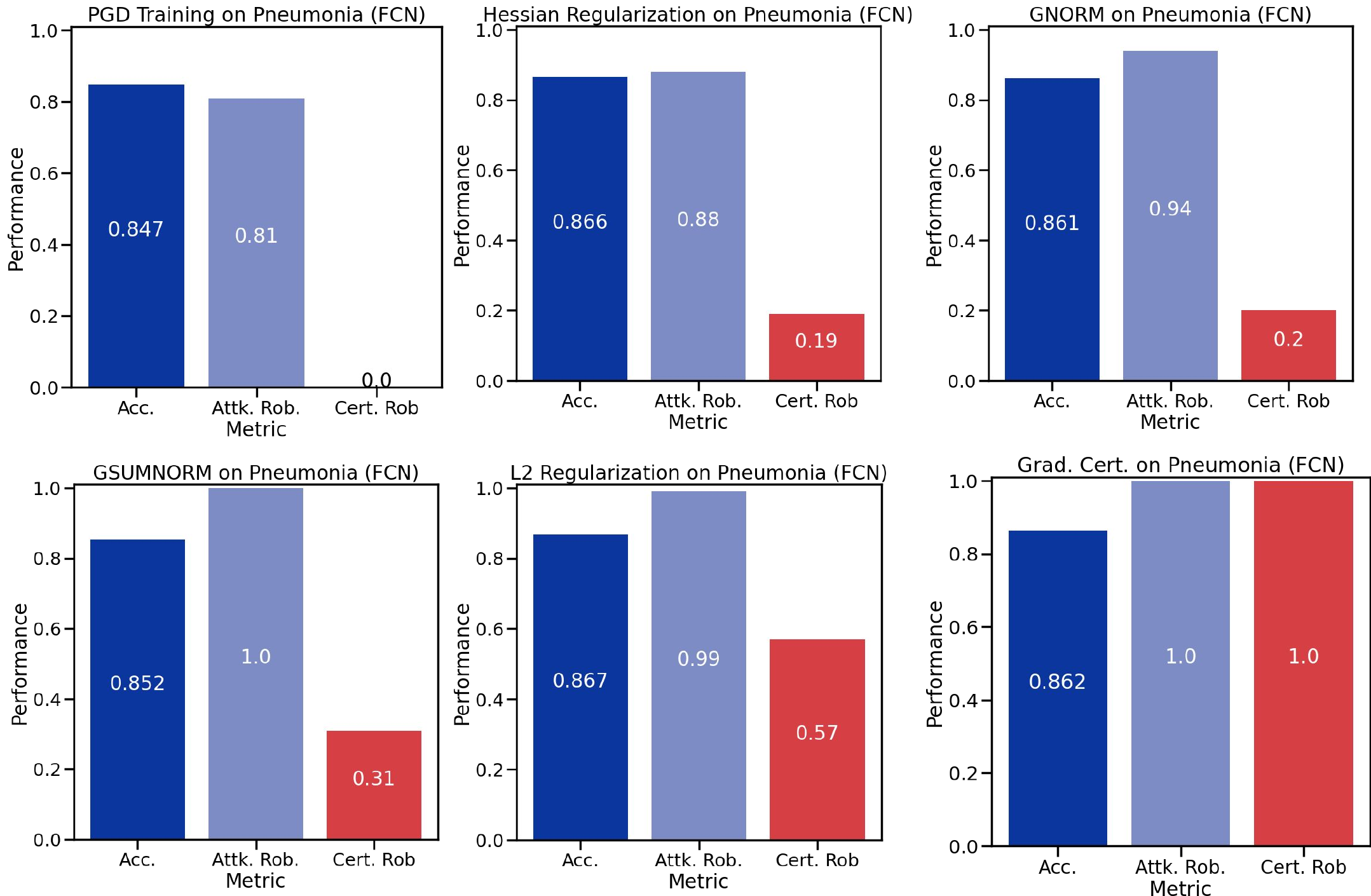}
    \caption{Our method out performs other robust explanation regularization approaches on the Pneumonia MedMNIST dataset. On the far left we plot standard training, center left we plot Hessian regularization, center right we plot L2 regularization, and far right we plot our method. We find that though regularization methods improve robustness against attacks and even provide some non-trivial certification in the case of L2 regularization, our method considerably out-performs each method with limited accuracy penalty.}
    \label{fig:pneumonia-compare}
\end{figure}

\section{\color{black}{Extended Comparisons}}\label{appendix:extendedcomparison}

In this Section, we provide further experimental comparisons between our method and previous regularization methods. We start by stating the optimization objective of each approach followed by the hyper-parameters used to generate the comparison in Figures~\ref{fig:mnist-compare} and \ref{fig:pneumonia-compare}. We conclude the section with a discussion of the results. The first regularization tested comes from \citep{drucker1992improving} and is also discussed in \citep{chen2019robust}. The authors propose an objective: 
$$ \mathcal{L}_{\text{L2}} = \mathcal{L}(f^{\theta}(x), y) + \alpha ||\nabla_x f^{\theta}(x') ||_2^2 $$
where we have added the additional term $x' = x + \mathcal{N}(0, \epsilon)$ and $\epsilon \in \mathbb{R}^{n}$ defines the width of $T$. This noise allows us to approximately minimize the gradient magnitude in a ball around the original input. We call this the L2 loss and is what is tested in the main text. All methods trained with this use $\alpha = 0.5$ with $\epsilon = 0.025$ for MNIST and $0.02$ for MEDMNIST. In \cite{chen2019robust} the authors propose an adversarial version of this loss. In this work, we generalize from integrated gradients to standard gradients and thus drop the `I' initial in the method name: 
$$ \mathcal{L}_{\text{GNORM}} = \mathcal{L}(f^{\theta}(x), y) + \alpha \max_{x' \in T} ||\nabla_x f^{\theta}(x) -  \nabla_{x'} f^{\theta}(x')||_1 $$
it is easy to observe that this is a similar loss to the Gradient Certified loss proposed in this paper save they opt for an $\ell_1$ norm penalty and approximately solve the optimization term via stochastic gradient descent whereas the gradient certified loss uses an $\ell_2$ norm penalty and computes an over-approximate worst-case solution to the optimization problem. For MNIST we use $\alpha = 1.0$, $\epsilon = 0.025$ and use $\alpha = 0.5$, $\epsilon = 0.02$ for MEDMNIST as these were found to be the best performing parameters. We use 10 iterations of projected gradient descent to solve the minimization term. We also highlight that as this optimization relies on the Hessian we must swap rectified linear units for softplus activations. We also consider \cite{madry2018towards} as a baseline method. The robust optimization proposes the following optimization objective: 
$$ \mathcal{L}_{\text{PGD}} = \max_{x' \in T} \mathcal{L}(f^{\theta}(x'), y) $$
where the maximization step is solved via projected gradient descent. When training NNs with this loss we again use 10 iterations of PGD with no restarts. We use $\epsilon = 0.025$ for MNIST and $\epsilon = 0.05$ for MEDMNIST. Finally, in \citep{chen2019robust} the authors propose combining the two yielding G-SUM-NORM:
$$ \mathcal{L}_{\text{GSUMNORM}} = \max_{x' \in T} \big[ \mathcal{L}(f^{\theta}(x), y) + ||\nabla_x f^{\theta}(x) -  \nabla_{x'} f^{\theta}(x')||_1 \big]$$
This is combination of robust optimization and robust gradient regularization. For this we use the same parameters as GNORM, i.e., $\alpha = 1.0$, $\epsilon = 0.025$ for MNIST and $\alpha = 0.5$, $\epsilon = 0.02$ for MEDMNIST.

In Figures~\ref{fig:mnist-compare} and \ref{fig:pneumonia-compare} we plot test set accuracy, attack robustness, and certified robustness, as defined for explanations in the main text, for each of the above methods. In Figure~\ref{fig:mnist-compare} we train the same one hidden layer neural network with 128 hidden neurons varying only the train time regularization used for each networks. We evaluate the robustness of these networks against an adversary with $\epsilon = 0.0125$ which is considerably lower than what is tested in the main text. This is to display non-trivial behavior for each baseline method. We find that all tested regularization methods lead to a non-trivial level attack robustness. Additionally we find that L2 regularization leads to the best input-certified explanations against an an adversary who perturbs the input by $\epsilon=0.0125$. We highlight that, as in the main text, Grad. Cert. considerably out-performs all other regularization methods in terms of certified performance. In Figure~\ref{fig:pneumonia-compare}, we again empirically compare the accuracy, attack robustness, and certified robustness of different regularization methods, this time on the pneumonia dataset. We evaluate the network using a smaller $\epsilon$ than is considered in the text, here using $\epsilon = 0.02$ with $\gamma = 0.0$ in order to distinguish methods that have comparably less robustness than offered by our method. Our empirical findings in Figure~\ref{fig:pneumonia-compare} mirror those of Figure~\ref{fig:mnist-compare}. In particular, we find that L2 regularization leads to the second best gradient performance while Grad. Cert. continues to out-perform all other methods. 

\section{{\color{black}Certification of Cosine Similarity}}\label{appendix:cosine}

In the main text we pose certification for targeted and untargetted attacks through the similarity function $h$ and focused on the $\ell_2$-norm as the $h$ function. In this section, we provide the details on the $h$ function and its approximation for when the measure of similarity (or dissimilarity) between two explanations is taken as the cosine similarity.

\subsection{\color{black}{Cosine Similarity Bound}}

We first recall that if $h(v, v')$ is the cosine similarity between the explanations $v$ and $v'$, then the function can be expressed as $$h(v, v') = \dfrac{\sum_{i=0}^{n} v'_i v_i}{\sqrt{\sum_{i=0}^{n} v'_i} \sqrt{\sum_{i=0}^{n} v_i}}$$

We can again propose values of $v^{cert}$ that allow us to certify if an adversarial example exists in a given input region. 

\textit{Targeted Attacks} Where $v^{targ}$ is the target explanation vector and $[v^{L}, v^{U}]$ is the interval computed by out method, outlined in Section~\ref{sec:computations}, we can compute values for $v' \in [v^{L}, v^{U}]$ that over-approximate the how close an adversary can get to $v^{targ}$ w.r.t. the cosine similarity. To minimize the cosine similarity, we would like to maximize the denominator while minimizing the numerator. As we only seek an over-approximate solution, we need not find a single value $v' \in [v^{L}, v^{U}]$. Instead we take $v^{denom}$ to be the smallest magnitude value in the range $[v^{L}, v^{U}]$ (either 0 or one of the end points) and we take $v^{numer}$ to be the value that minimizes the dot product with the target vector (again, either 0 or one of the end points). The resulting value is an over-approximate minimum for the smallest cosine similarity between the vector $v^{targ}$ and any vector in the interval $[v^{L}, v^{U}]$.


\section{\color{black}{Proof and Theoretical Discussion}}\label{appendix:proofs}

In this section of the Appendix we provide a formal proof of Lemma 1 as well as a discussion of its tightness.

\subsection{\color{black}{Proof of Lemma 1}}

Recall that Lemma 1 operates on intervals over matrices. We take the first matrix interval to be $[A^L, A^U]$ and the second matrix interval to be $[B^L, B^U]$. Denoting generic operands in the interval with $A \in \mathbb{R}^{n \times m}$ and $B \in \mathbb{R}^{m \times k}$ we would like to find upper and lower bounds on the product of any two matrices from these intervals (the resulting product being a matrix in $\mathbb{R}^{n \times k}$). For any $i \in [n]$ and $j \in [k]$ the $i,j$ entry in the product-space can be written as $A_{i:} \cdot B_{:j}$, the dot-product of the $i^{th}$ row of $A$ with the $j^{th}$ column of $B$, which is defined as $\sum_{t=0}^{m} A_{i, t}B_{t,j}$. We now focus bounding the maximum of this dot product (the logic for the minimum follows the same pattern). This is maximized when each term $A_{i, t}B_{t,j}$ is maximized. Given the bi-linear nature of this optimization, the maximum is obtained at one of the four corners of the rectangle $[A^{L}_{i, t},A^{U}_{i, t}] \times [B^{L}_{t,j}, B^{U}_{t,j}]$. We now show that Lemma 1 over-approximates the maximizing corner. Recall our upper bound on this rectangle is given as is $A^\mu_{i,t} B^\mu_{t,j} + |A^\mu_{i,t}|B^r_{t,j} + A^r_{i,t} |B^\mu_{t,j}| + |A^r_{i,t}||B^r_{t,j}|$. 

In the case that $A^L = A^U$ and $B^L = B^U$, then we have that $A^L = A^U = A^{\mu}$ and $B^L = B^U = B^{\mu}$. Thus all of the corners of the rectangle ($[A^{L}_{i, t},A^{U}_{i, t}] \times [B^{L}_{t,j}, B^{U}_{t,j}]$) are equal, and the bound returns this value exactly, $A^{\mu}_{i, t} B^{\mu}_{t,j}$, as all values super-scripted $r$ are equal to 0.

In the case that $A^L \neq A^U$ but $B^L = B^U$, then we have that $B^L = B^U = B^\mu$, thus the maximum of the rectangle either occurs at $A^{L}_{i,t}B^\mu_{t,j}$ or $A^{U}_{i,t}B^\mu_{t,j}$ with the other containing the minimum. The center of the interval by definition is $A^{\mu}_{i,t}B^\mu_{t,j}$ and we observe that the width of the interval between the maximum and minimum (regardless of which is which) is $2A^r_{i,t} |B^{\mu}_{t,j}|$, thus the maximum is obtained at $A^{\mu}_{i,t}B^\mu_{t,j} + A^r |B^{\mu}|$ and the minimum at $A^{\mu}_{i,t}B^\mu_{t,j} - A^r |B^{\mu}|$, as prescribed by the bound (as $B^r_{t,j} = 0$ erasing the contribution of the other terms).

Finally we have the case in which $A^L \neq A^U$ and $B^L \neq B^U$. In this case we have that the maximum and minimum could occur at any one of the four corners of the rectangle. As before, the center of the interval in product space is $A^{\mu}_{i,t}B^\mu_{t,j}$. And, assuming that $B^\mu_{t,j} \neq 0$, the width contributed by the interval $[A^{L}_{i,t}, A^{U}_{i,t}]$ is given by $A^r_{i,t} |B^\mu_{t,j}|$. Equivalently, assuming that $A^\mu_{i,t} \neq 0$, the width contributed by the interval $[B^{L}_{i,t}, B^{U}_{i,t}]$ is given by $|A^\mu_{i,t}|B^r_{t,j}$. In the case that both $A^\mu_{i,t} = 0$  and $B^\mu_{t,j} = 0$, the term $A^r_{i,t} B^r_{t,j}$ jointly accounts for both widths exactly. In the case that $A^\mu_{i,t} = 0$ and $B^\mu_{t,j} \neq 0$ our bound introduces approximation error by over counting the width of $[B^U_{t,j}, B^L_{t,j}]$. 
However this approximation is sound as we have over-approximated the minimum or maximum.

Above we have shown that Lemma 1 is a sounds over-approximation by exhausting all of the possible interval configurations and showing that Lemma 1 is sound for each. 

\subsection{\color{black}{Discussion of Lemma 1 Tightness}}

The bound provided in Lemma 1 represents interval bound propagation jointly over two matrices. As an interval bound, it is exact in every case save for when one of the bounds is centered exactly at zero and the other is not. In this case, the alternative (tighter) interval bounding procedure would be to compute each of the four corners $[A^{L}_{i, t},A^{U}_{i, t}] \times [B^{L}_{t,j}, B^{U}_{t,j}]$ and subsequently taking the maximum and minimum. This procedure 
requires an element-wise maximum and minimum operation which make the optimization of such a bound more challenging for auto-differentiation software (Pytorch, Tensorflow, etc), thus Lemma 1 is considerably more desirable when one wants to incorporate our bounds into training. Though at test-time one can gain marginal improvements in certification by taking elementwise maximums and minimums.

\section{{\color{black}Further Related Work}}\label{appendix:futherlit}

In our literature review we focus on gradient-based explanations and contextualizing the study of their robustness. Here, we briefly cover explanation methods that use robustness as a primary desiderata. In \citep{ignatiev2019abduction} the authors rely on abductive reasoning to get guarantees that are guaranteed to be \textit{minimal} (e.g., in the number of explaining features used) but are not guaranteed to be robust. The authors of \citep{la2021guaranteed} build on abductive-based explanations and consider explanations that are both minimal and optimally robust. In \citep{blanc2021provably} the authors also consider provable robustness as a key desiderata and achieve this by sampling a black-box model exponentially many times and thus deriving statistical guarantees. Similarly an iterative, greedy strategy is employed by \citep{ribeiro2018anchors} to get some statistical guarantees on the quality of their explanations. In counterfactual explanations, \citep{blanc2022query} theoretically guarantees optimal counterfactuals with their algorithm with analysis of their query complexity. In \citep{mohammadi2021scaling}, the authors provide provable guarantees in terms of optimal distance, i.e., nearest explanation, and perfect coverage.


\end{document}